\pdfoutput=1

\documentclass{article}
\usepackage{colm2024_conference}

\usepackage{url}
\usepackage{booktabs}
\usepackage{hyperref}
\usepackage{graphicx}
\usepackage{microtype}
\usepackage{multicol}
\usepackage{xcolor}   
\usepackage{multirow}
\usepackage{lipsum}
\usepackage{enumitem}
\usepackage{pifont}
\usepackage{colortbl}
\usepackage{tabularx}
\usepackage{caption} 
\usepackage{newfloat}
\usepackage{amsmath,amsfonts,amssymb,bbm}
\usepackage{amsthm}

\DeclareFloatingEnvironment[
  fileext=loa,
  listname={List of Algorithms},
  name=Algorithm,
  placement=tbp
]{algorithm}

\title{FlashQLA: Efficient Context Parallelism for Gated DeltaNet}

\author{
\vspace{1.5mm}
Xi Lin$^{1,2,*}$ \quad
Chengruidong Zhang$^{2,*}$ \quad
Huiqiang Jiang$^{2}$ \quad
Xiao Li$^{2}$ \quad
Zekun Wang$^{2}$ \\
Junda Feng$^{2}$ \quad
Yunfei Mao$^{2}$ \quad
Siqi Zhang$^{2}$ \quad
Siyu Wang$^{2}$ \quad
Man Yuan$^{2,\dagger}$ \quad
Rui Men$^{2,\dagger}$ \\
Bo Zheng$^{2,\dagger}$ \quad
Dayiheng Liu$^{2,\dagger}$ \quad
Bohan Zhuang$^{1,\dagger}$ \quad
Jianwei Zhang$^{2,\dagger}$
\\[1.0em]
$^{1}$Zhejiang University \qquad
$^{2}$Alibaba Token Hub, Alibaba Group
\\[0.4em]
{\small
$^{*}$Equal contribution.
\qquad
$^{\dagger}$Corresponding author.
}
}

\begin{document}

\maketitle

\begin{abstract}
Gated DeltaNet (GDN) achieves linear computational complexity through recurrent state updates, but sequential dependencies between chunks limit parallelism, particularly for long-context workloads with small batch sizes or limited head-level parallelism per GPU. We present FlashQLA, a context-parallel execution framework for GDN that exposes additional sequence-level parallelism while reducing boundary-state preprocessing overhead. FlashQLA reduces boundary-state preprocessing overhead through decay-aware preparation and extends sequence parallelism across GPUs with Dual CP, which shares summary computation across parallelism levels. Workload-adaptive Auto CP and CP-aware kernel co-design further improve execution efficiency. FlashQLA achieves $1.7\times$--$3.1\times$ forward speedup and up to $4.4\times$ backward speedup over Flash Linear Attention (FLA) on a single GPU. On eight GPUs with context lengths up to 1M tokens, Dual CP achieves up to $4.8\times$ forward speedup and $4.4\times$ speedup for combined forward and backward execution over FLA KCP. Code is available at \url{https://github.com/QwenLM/FlashQLA}.
\end{abstract}

\section{Introduction}
\label{sec:intro}

Linear attention replaces the quadratic token-wise interactions of softmax attention with linear-complexity computation through recurrent state updates~\citep{katharopoulos2020transformers}. Gated DeltaNet (GDN) further combines delta-rule updates with data-dependent state decay, allowing the model to selectively update the recurrent state while controlling how much historical information is retained \citep{yang2024parallelizing,yang2025gated}. GDN has been adopted in large-scale hybrid language models, from Qwen3-Next to the Qwen3.8 series and Qwen3.8-Flash-Next, highlighting its growing importance for efficient long-context modeling~\citep{qwen3next,qwen3.8,qwen2026design}.

Efficient GDN prefill execution relies on a chunk-wise formulation based on the extended WY representation, which enables parallel computation within each chunk using GPU-efficient matrix multiplications~\citep{yang2025gated}. However, the recurrent state must still be propagated sequentially across chunks, creating a long dependency chain as sequence length increases. This inter-chunk recurrence becomes a major performance bottleneck for long-context execution.

When batch size and head count provide insufficient parallelism, the sequence dimension offers an additional source of concurrency. Context parallelism (CP) exploits this opportunity by partitioning the sequence to enable parallel execution. However, each partition depends on the states produced by preceding partitions. Kimi Context Parallel(KCP)~\citep{kimi2026k3} resolve these dependencies through boundary-state preprocessing, consisting of state-summary preparation to compute local summaries and boundary-state correction to reconstruct partition input states. This preprocessing, however, can itself become a major bottleneck. For example, it accounts for 50.8\% of the forward execution time of KCP in Flash Linear Attention(FLA)~\citep{yang2024fla} on eight GPUs with a 1M-token sequence, limiting the performance gains achievable through context parallelism.

We present \emph{FlashQLA}, an efficient context-parallel execution framework for GDN that exposes sequence-level parallelism while reducing its preprocessing overhead. Conventional state-summary preparation processes every chunk within a partition, regardless of how strongly historical state contributions are attenuated by gate decay. FlashQLA exploits this head-wise decay behavior through decay-aware preparation, truncating computation for strongly decaying heads while retaining full-history preparation for heads with longer-range dependencies. This avoids unnecessary computation and substantially reduces preprocessing overhead while maintaining sufficient numerical accuracy in practice.

Increasing the number of partitions exposes more parallelism but also increases preprocessing and scheduling overhead. FlashQLA therefore introduces \emph{Auto CP}, which uses a lightweight latency proxy model and wave-aware search to determine whether intra-card CP is beneficial and select an efficient partition length for each workload.

Efficient GDN execution at longer context lengths also benefits from extending sequence-level parallelism across multiple GPUs. FlashQLA therefore introduces \emph{Dual CP}, which jointly combines intra-card and inter-card context parallelism. Naively combining these two levels would duplicate state-summary preparation. Dual CP avoids this redundancy by sharing summary computation across both levels and performing hierarchical boundary-state correction. avoiding redundant computation while increasing sequence-level parallelism within each GPU.

Across both intra- and inter-card execution, FlashQLA further co-designs its GPU kernels with the CP execution schedule, incorporating algebraic simplification, CP-aware fusion, and manual warp specialization to improve hardware utilization and execution efficiency.

FlashQLA achieves $1.7\times$--$3.1\times$ forward speedup and up to $4.4\times$ backward speedup over FLA on a single GPU. Scaling to eight GPUs with context lengths up to 1M tokens, Dual CP achieves up to $4.8\times$ forward speedup and $4.4\times$ speedup for combined forward and backward execution over FLA KCP.

This report makes the following contributions:
\begin{enumerate}[leftmargin=*,itemsep=2pt,topsep=3pt]
    \item We introduce decay-aware preparation, which exploits head-wise gate decay to substantially reduce the boundary-state preprocessing overhead of context parallelism.

    \item We develop \emph{Auto CP}, a workload-adaptive partitioning strategy that accounts for the trade-off between increased sequence-level parallelism and the overhead of finer-grained partitioning, enabling effective intra-card context parallelism across diverse workloads.

    \item We propose \emph{Dual CP}, which efficiently integrates intra-card and inter-card context parallelism through shared summary computation and hierarchical boundary-state correction, avoiding duplicated computation across parallelism levels.

    \item We co-design GPU kernels with the CP execution schedule through algebraic simplification, stage-aware fusion, and manual warp specialization. Together, these techniques enable FlashQLA to achieve $1.7\times$--$3.1\times$ forward speedup over FLA on a single GPU and up to $4.8\times$ over FLA KCP on eight GPUs with context lengths up to 1M tokens.
\end{enumerate}

\section{Background and Problem Formulation}
\label{sec:background}

\subsection{Gated DeltaNet}
\label{sec:gdn-review}

Gated DeltaNet~\citep{yang2025gated} maintains a recurrent state that is updated at each token by combining a data-dependent decay with a delta-rule correction. Let
$\mathbf{S}_t\in\mathbb{R}^{d_k\times d_v}$ denote the recurrent state after token $t$, with
$\mathbf{q}_t,\mathbf{k}_t\in\mathbb{R}^{1\times d_k}$,
$\mathbf{v}_t\in\mathbb{R}^{1\times d_v}$,
decay gate $\alpha_t\in(0,1)$, and delta-update strength $\beta_t\in(0,1)$.
The gated delta rule is
\begin{equation}
\mathbf{S}_t
=
\alpha_t
\left(\mathbf{I}-\beta_t\mathbf{k}_t^\top\mathbf{k}_t\right)\mathbf{S}_{t-1}
+
\beta_t\mathbf{k}_t^\top\mathbf{v}_t,
\qquad
\mathbf{o}_t=\mathbf{q}_t\mathbf{S}_t.
\label{eq:gdn-token}
\end{equation}
The decay gate $\alpha_t$ controls the retention of the previous state, while $\beta_t$ determines the strength of the delta-rule update along the current key direction.

The token-wise recurrence is inherently sequential along the sequence dimension. To expose parallelism, GDN groups $C$ consecutive tokens into a chunk and uses an extended WY representation to parallelize updates within each chunk~\citep{yang2025gated}. For simplicity, we assume full chunks of size \(C\); partial chunks can be handled by padding or masking.
We use bracketed subscripts $[c]$ to index chunks, distinguishing them from token-level subscripts.
Let
$\mathbf{Q}_{[c]},\mathbf{K}_{[c]}\in\mathbb{R}^{C\times d_k}$ and
$\mathbf{V}_{[c]}\in\mathbb{R}^{C\times d_v}$
denote the queries, keys, and values of chunk $c$, and let
$\alpha_{[c],t}$ and $\beta_{[c],t}$ denote the decay gate and update strength
at local token position $t=1,\ldots,C$.
We define
\begin{equation}
\gamma_{[c],t}
:=
\prod_{j=1}^{t}\alpha_{[c],j},
\qquad
\mathbf{B}_{[c]}
:=
\operatorname{diag}
\left(
\beta_{[c],1},\ldots,\beta_{[c],C}
\right),
\qquad
\boldsymbol{\Gamma}_{[c]}
:=
\operatorname{diag}
\left(
\gamma_{[c],1},\ldots,\gamma_{[c],C}
\right).
\label{eq:gdn-bgamma}
\end{equation}

Let $\mathbf{S}_{[c]}$ denote the recurrent state after processing the first $c$ chunks, with $\mathbf{S}_{[0]}=\mathbf{S}_0$.
The chunk-wise GDN formulation is
\begin{align}
\mathbf{A}_{[c]}
&=
\left(
\mathbf{I}+
\operatorname{StrictLower}\!\left(
\mathbf{B}_{[c]}\boldsymbol{\Gamma}_{[c]}
\mathbf{K}_{[c]}\mathbf{K}_{[c]}^\top
\boldsymbol{\Gamma}_{[c]}^{-1}
\right)
\right)^{-1},
\label{eq:gdn-chunk-a}\\
\mathbf{U}_{[c]}
&=
\mathbf{A}_{[c]}\mathbf{B}_{[c]}\mathbf{V}_{[c]},
\label{eq:gdn-chunk-u}\\
\mathbf{W}_{[c]}
&=
\mathbf{A}_{[c]}\mathbf{B}_{[c]}
\boldsymbol{\Gamma}_{[c]}\mathbf{K}_{[c]},
\label{eq:gdn-chunk-w}\\
\mathbf{S}_{[c]}
&=
\gamma_{[c],C}\mathbf{S}_{[c-1]}
+
\gamma_{[c],C}\mathbf{K}_{[c]}^\top
\boldsymbol{\Gamma}_{[c]}^{-1}
\left(
\mathbf{U}_{[c]}-\mathbf{W}_{[c]}\mathbf{S}_{[c-1]}
\right),
\label{eq:gdn-chunk-state}\\
\mathbf{O}_{[c]}
&=
\boldsymbol{\Gamma}_{[c]}\mathbf{Q}_{[c]}\mathbf{S}_{[c-1]}
+
\boldsymbol{\Gamma}_{[c]}
\operatorname{Lower}
\left(
\mathbf{Q}_{[c]}\mathbf{K}_{[c]}^\top
\right)
\boldsymbol{\Gamma}_{[c]}^{-1}
\left(
\mathbf{U}_{[c]}-\mathbf{W}_{[c]}\mathbf{S}_{[c-1]}
\right).
\label{eq:gdn-chunk-output}
\end{align}

Here, $\mathbf{A}_{[c]}$, $\mathbf{U}_{[c]}$, and $\mathbf{W}_{[c]}$ are chunk-local auxiliary quantities, while
$\operatorname{StrictLower}$ and $\operatorname{Lower}$ denote the strictly lower- and lower-triangular parts of a matrix, respectively.

FLA~\citep{yang2024fla} implements chunk-wise GDN through multiple GPU kernel stages. In the FLA implementation used as our baseline, its forward execution constructs the extended WY representation through partially fused intra-chunk operations, followed by separate kernels for recurrent-state propagation and output computation. This multi-stage design requires materializing intermediate tensors between stages, introducing additional kernel launches and global memory accesses.

Although the extended WY representation parallelizes computation within each chunk, the chunk boundary states remain sequential: $\mathbf{S}_{[c+1]}$ depends on $\mathbf{S}_{[c]}$. Consequently, chunks within the same sequence cannot be processed independently without first resolving their boundary-state dependencies, motivating context-parallel execution.

\subsection{Boundary-State Preprocessing for GDN Context Parallelism}
\label{sec:cp-theory}

To apply context parallelism to GDN, the chunk sequence is partitioned into subsequences, but each partition's input state depends on preceding chunks. KCP~\citep{kimi2026k3} address this dependency by first constructing the local state transition of each partition and then determining the corresponding boundary states. This boundary-state preprocessing consists of two stages: state-summary preparation computes a zero-initialized terminal state and a cumulative transition for each partition, while boundary-state correction determines the input state of each partition before parallel execution.

The chunk-wise state update in Eq.~\eqref{eq:gdn-chunk-state} is affine in its input state:
\begin{equation}
\begin{aligned}
\mathbf{S}_{[c]}
&=
\mathbf{M}_{[c]}\mathbf{S}_{[c-1]}
+
\mathbf{E}_{[c]},\\
\mathbf{M}_{[c]}
&:=
\gamma_{[c],C}
\left(
\mathbf{I}
-
\mathbf{K}_{[c]}^\top
\boldsymbol{\Gamma}_{[c]}^{-1}
\mathbf{W}_{[c]}
\right),\\
\mathbf{E}_{[c]}
&:=
\gamma_{[c],C}
\mathbf{K}_{[c]}^\top
\boldsymbol{\Gamma}_{[c]}^{-1}
\mathbf{U}_{[c]}.
\end{aligned}
\label{eq:chunk-summary}
\end{equation}
Here, $\mathbf{M}_{[c]}$ propagates the incoming state through chunk $c$, while $\mathbf{E}_{[c]}$ represents the chunk-local contribution independent of that state.

An important property of this affine recurrence is that the difference between two state trajectories evolves independently of the additive contributions. Let $\widetilde{\mathbf{S}}_{[c]}$ denote the state obtained from a zero-initialized trajectory with the same chunk inputs, and define $\Delta\mathbf{S}_{[c]}:=\mathbf{S}_{[c]}-\widetilde{\mathbf{S}}_{[c]}$.
Their difference after processing chunks $a$ through $b$ satisfies
\begin{equation}
\Delta\mathbf{S}_{[b]}
=
\mathbf{M}_{[b]}
\cdots
\mathbf{M}_{[a]}
\Delta\mathbf{S}_{[a-1]}.
\label{eq:bg-state-difference}
\end{equation}
This property allows the terminal state under an arbitrary initial condition to be reconstructed from a zero-state rollout and the corresponding cumulative transition. A detailed derivation is provided in Appendix~\ref{app:boundary-state-preprocessing}.

Let the $L$ chunks be divided into $N$ ordered partitions, where partition $i$ contains $l_i$ consecutive chunks and $\sum_{i=1}^{N}l_i=L$.
We use superscripts $(i)$ to identify partitions and bracketed subscripts $[j]$ to index chunks locally within each partition. Let $\mathbf{S}_{[0]}^{(i)}$ denote the input state of partition $i$, and $\mathbf{S}_{[j]}^{(i)}$ the state after processing its first $j$ chunks. The input boundary states satisfy $\mathbf{S}_{[0]}^{(1)}=\mathbf{S}_0$ and $\mathbf{S}_{[0]}^{(i)}=\mathbf{S}_{[l_{i-1}]}^{(i-1)}$ for $i>1$.

During state-summary preparation, each partition independently performs a zero-state rollout, producing $\widetilde{\mathbf{S}}_{[l_i]}^{(i)}$ with $\widetilde{\mathbf{S}}_{[0]}^{(i)}=\mathbf{0}$, together with the cumulative transition
\begin{equation}
\mathbf{M}_{[l_i\leftarrow1]}^{(i)}
:=
\mathbf{M}_{[l_i]}^{(i)}
\cdots
\mathbf{M}_{[1]}^{(i)}.
\label{eq:bg-cumulative-transition}
\end{equation}

Applying Eq.~\eqref{eq:bg-state-difference}, the final state of partition $i$ is
\begin{equation}
\mathbf{S}_{[l_i]}^{(i)}
=
\widetilde{\mathbf{S}}_{[l_i]}^{(i)}
+
\mathbf{M}_{[l_i\leftarrow1]}^{(i)}
\mathbf{S}_{[0]}^{(i)}.
\label{eq:bg-forward-boundary-correction}
\end{equation}

State-summary preparation therefore proceeds in parallel across partitions, with each partition computing two summaries, $\widetilde{\mathbf{S}}_{[l_i]}^{(i)}$ and $\mathbf{M}_{[l_i\leftarrow1]}^{(i)}$. Starting from $\mathbf{S}_{[0]}^{(1)}=\mathbf{S}_0$, boundary-state correction applies Eq.~\eqref{eq:bg-forward-boundary-correction} sequentially across partition summaries to determine the input states of subsequent partitions. Once preprocessing is complete, main computation of all partitions can proceed in parallel.

Although context parallelism exposes additional sequence-level parallelism for the main computation, it introduces substantial boundary-state preprocessing overhead. In particular, state-summary preparation requires an additional sequential recurrence over the chunks within each partition, with computational cost comparable to the main forward pass. This overhead can significantly diminish the performance gains from parallel execution, regardless of whether partitions are distributed within or across GPUs. On eight GPUs with a 1M-token sequence, boundary-state preprocessing accounts for $50.8\%$ of FLA KCP's forward execution time, highlighting preprocessing overhead as a critical bottleneck for efficient context-parallel execution.

\section{Decay-Aware Preparation}
\label{sec:decay-aware-cp}

Conventional boundary-state preprocessing, as described in Section~\ref{sec:cp-theory}, requires state-summary preparation over all chunks within each partition. FlashQLA exploits head-wise variation in gate decay to reduce this cost, using truncated preparation for strong-decay heads while retaining full-history preparation for heads with longer-range dependencies.

\subsection{Head-Wise Gate Decay Characteristics}
\label{sec:gate-distribution}

We first examine how rapidly historical information decays across GDN heads. Given the chunk-level cumulative decay factor $\gamma_{[c],C,h}$ for head $h$, we define
\begin{align}
D_h(c-k+1,c)
&:=
\prod_{r=c-k+1}^{c}
\gamma_{[r],C,h},
\label{eq:cumulative-gate-decay}\\
K_{c,h}
&:=
\min\left\{
1\le k \le c:
D_h(c-k+1,c)<\mathrm{TOL}
\right\},
\label{eq:decay-window-definition}\\
W_{c,h}
&:= C K_{c,h},
\label{eq:decay-window-tokens}
\end{align}
where $\mathrm{TOL}$ is a threshold for cumulative decay, $K_{c,h}$ is the decay window in chunks, and $W_{c,h}$ is the corresponding window in tokens.  A smaller decay window indicates faster attenuation of historical information. If no window satisfies the threshold within the available history, we set $K_{c,h}=\infty$ and use full-history preparation.

\begin{figure}[t]
    \centering
    \includegraphics[width=0.9\linewidth]{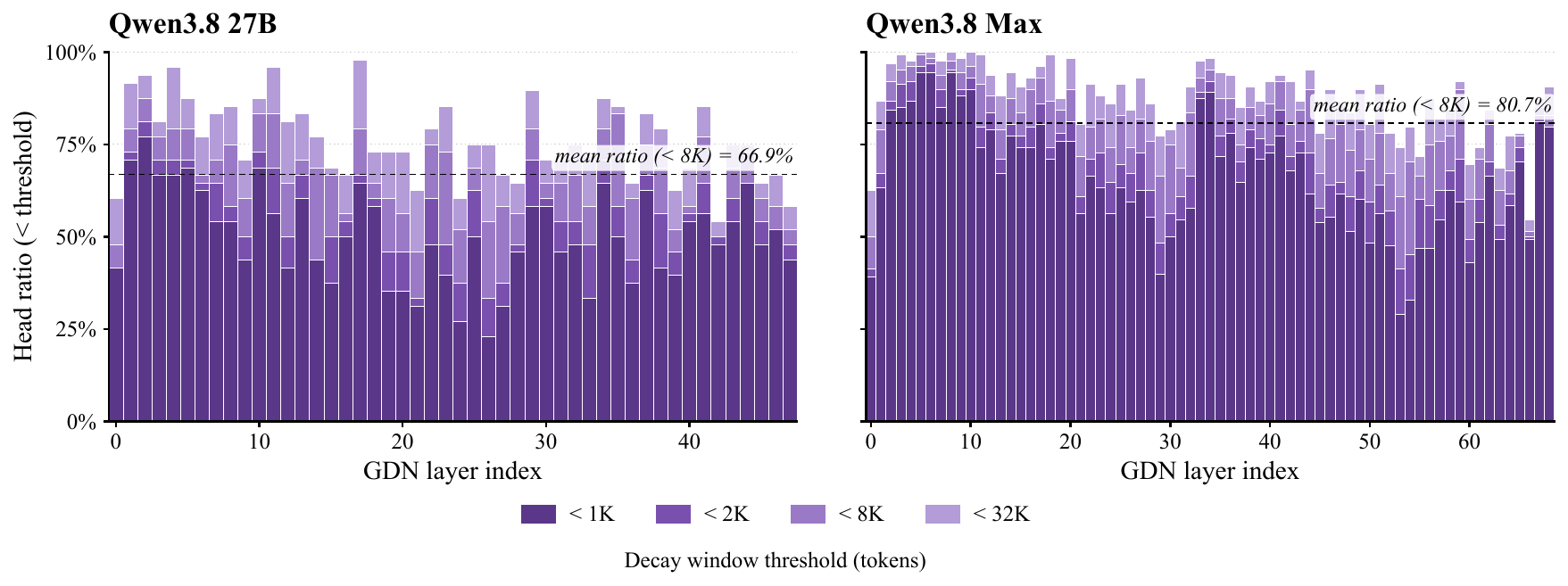}
    \caption{
    \textbf{Head-wise decay window distribution.}
    Fraction of heads with decay windows shorter than 1K, 2K, 8K, and 32K tokens across layers in Qwen3.8-27B and Qwen3.8-2.4T-A95B, evaluated on 32K-token sequences with $\mathrm{TOL}=e^{-10}$.
    Overall, 66.9\% and 80.7\% of heads have decay windows shorter than 8K tokens in the two models, respectively, highlighting the prevalence of short decay windows.
    }
    \label{fig:decay-window-distribution}
\end{figure}

To characterize the prevalence of short decay windows, we evaluate Qwen3.8-27B and Qwen3.8-2.4T-A95B~\citep{qwen3.8} on 32K-token sequences with $\mathrm{TOL}=e^{-10}$, measuring the fraction of heads with decay windows shorter than 1K, 2K, 8K, and 32K tokens in each layer. As shown in Figure~\ref{fig:decay-window-distribution}, short decay windows are prevalent across both models and their layers. In particular, 66.9\% of heads in Qwen3.8-27B and 80.7\% in Qwen3.8-2.4T-A95B have decay windows shorter than 8K tokens. These results indicate that rapid gate decay is widespread in GDN, providing substantial opportunities to reduce boundary-state preparation overhead through decay-aware computation.

We further examine how the decay behavior of each head varies across inputs. For each model, we collect log-gate values from 24 sequences of 32K tokens and estimate each head's decay window from its median log-gate value, assuming a constant decay rate until the cumulative decay reaches $\mathrm{TOL}=e^{-10}$.

\begin{figure}[t]
    \centering
    \includegraphics[width=0.9\linewidth]{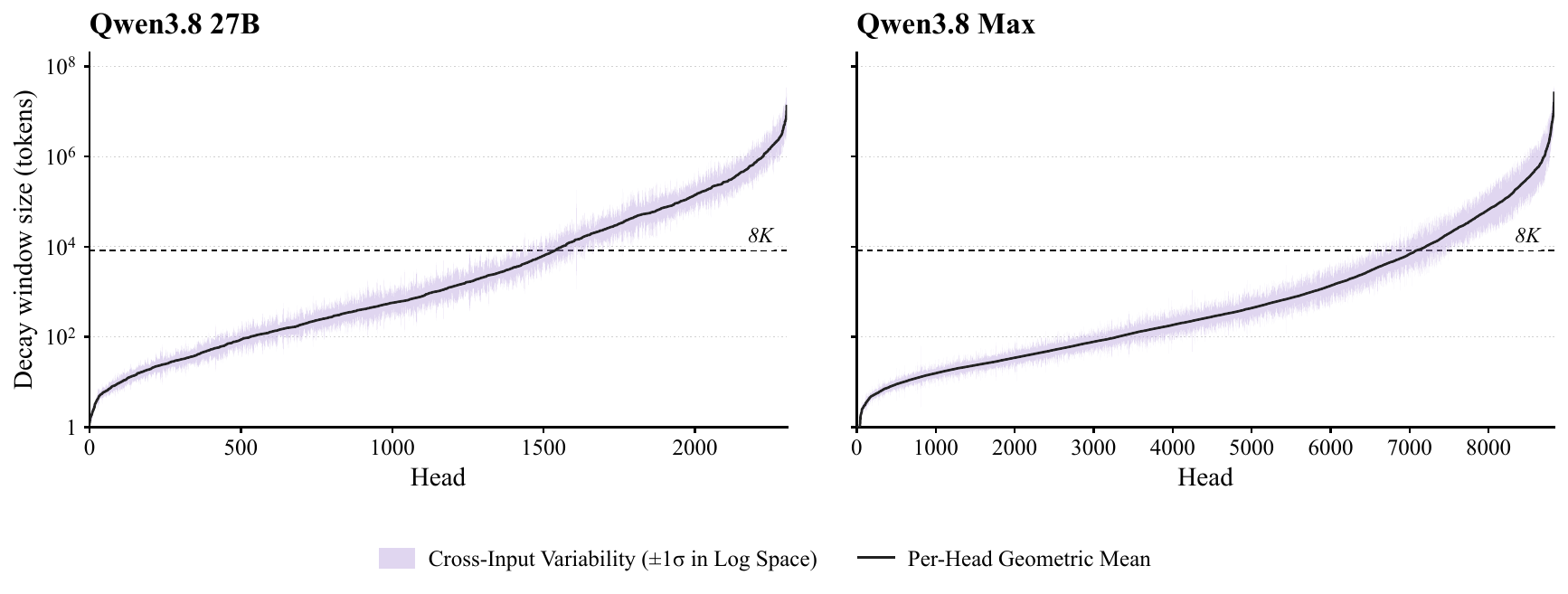}
    \caption{
    \textbf{Input dependence of head-wise decay windows.}
    Estimated decay windows across 24 inputs for Qwen3.8-27B and Qwen3.8-2.4T-A95B. Curves show geometric means, with shaded regions indicating $\pm 1$ standard deviation in log space. Windows beyond 32K are extrapolated from measured gate values.
    }
    \label{fig:decay-window-consistency}
\end{figure}

As shown in Figure~\ref{fig:decay-window-consistency}, heads exhibit characteristic decay patterns across inputs, while their exact window sizes remain input-dependent, motivating online window estimation.

These observations motivate adapting state-summary preparation to the input-dependent decay behavior of individual heads. We next examine how gate decay attenuates historical state contributions and enables truncated preparation.

\subsection{Gate Decay and Effective History}
\label{sec:gate-decay-observation}

Consider the boundary after chunk $b$ for head $h$, with decay window $K_{b,h}$ defined in Eq.~\eqref{eq:decay-window-definition}. To examine the effect of truncating historical computation, we retain only the most recent $K_{b,h}$ chunks, from $b-K_{b,h}+1$ to $b$, and initialize the state at the beginning of this window to zero. The truncated and full-history trajectories process identical chunk inputs but start from different initial states. We factorize the chunk transition in Eq.~\eqref{eq:chunk-summary} into a cumulative gate-decay factor and a content-dependent transformation:
\begin{equation}
\mathbf{M}_{[c]}
=
\gamma_{[c],C}\,
\overline{\mathbf{M}}_{[c]},
\qquad
\overline{\mathbf{M}}_{[c]}
:=
\mathbf{I}
-
\mathbf{K}_{[c]}^\top
\boldsymbol{\Gamma}_{[c]}^{-1}
\mathbf{W}_{[c]}.
\label{eq:decay-aware-transition-factorization}
\end{equation}

Let $\Delta\mathbf{S}_{[b-K_{b,h}]}$ denote the difference between the full-history and truncated-history states at the beginning of the retained window. By the state-difference propagation property in Eq.~\eqref{eq:bg-state-difference}, the difference at the boundary after chunk $b$ is
\begin{align}
\Delta\mathbf{S}_{[b]}
&=
\left(
\prod_{c=b-K_{b,h}}^{b}
\gamma_{[c],C}
\right)
\overline{\mathbf{M}}_{[b]}
\cdots
\overline{\mathbf{M}}_{[b-K_{b,h}]}
\Delta\mathbf{S}_{[b-K_{b,h}]},
\label{eq:decay-aware-state-difference}
\end{align}

This decomposition shows that cumulative gate decay attenuates the influence of historical states outside the retained window, motivating truncated state-summary preparation. FlashQLA uses a small threshold of $\mathrm{TOL}=e^{-10}$ to determine the truncation window.

\subsection{Decay-Aware State-Summary Preparation}
\label{sec:head-wise-preprocess}

Building on the gate-decay characteristics discussed above, FlashQLA adaptively selects the preparation window for each boundary and head. Since gate values are already available, the decay window $K_{c,h}$ can be computed online through cumulative sums of log-decay factors. We set the maximum truncation window $K_{\max}$ to the length of the preceding partition and classify each boundary-head pair as \emph{strong-decay} when $K_{c,h}\le K_{\max}$, and as \emph{weak-decay} otherwise.

For clarity, we suppress the head index and let $K_i:=K_{b_i,h}$ denote the decay window at the input boundary of partition $i$, where $b_i=\sum_{p=1}^{i-1}l_p$ is the global index of the preceding partition's last chunk. For a strong-decay boundary, FlashQLA performs a zero-state rollout over only the final $K_i$ chunks of the preceding partition, yielding the truncated state $\widetilde{\mathbf{S}}_{[l_{i-1}],K_i}^{(i-1)}$. The resulting approximate boundary state is
\begin{equation}
\widehat{\mathbf{S}}_{[0]}^{(i)}
=
\begin{cases}
\widetilde{\mathbf{S}}_{[l_{i-1}],K_i}^{(i-1)},
& K_i\le K_{\max},
\\[2mm]
\widetilde{\mathbf{S}}_{[l_{i-1}]}^{(i-1)}
+
\mathbf{M}_{[l_{i-1}\leftarrow 1]}^{(i-1)}
\widehat{\mathbf{S}}_{[0]}^{(i-1)},
& K_i>K_{\max},
\end{cases}
\qquad i>1,
\label{eq:decay-aware-boundary-assembly}
\end{equation}
with $\widehat{\mathbf{S}}_{[0]}^{(1)}=\mathbf{S}_0$.
For strong-decay boundaries, FlashQLA reduces preparation cost through truncated zero-state rollouts, while weak-decay boundaries retain full-history preparation to preserve longer-range dependencies.
The resulting $\widehat{\mathbf{S}}_{[0]}^{(i)}$ approximates the true boundary state $\mathbf{S}_{[0]}^{(i)}$ with limited numerical error in practice and serves as the initial state for the subsequent partition-parallel computation.

\section{Auto CP}
\label{sec:auto-cp}

The benefit of intra-card context parallelism depends strongly on the workload. Finer sequence partitioning exposes additional parallelism but also increases preprocessing and scheduling overhead. Consequently, neither always enabling context parallelism nor using a fixed partition size is optimal across different input configurations. FlashQLA therefore introduces \emph{Auto CP}, which uses a lightweight performance proxy model to adaptively determine whether and how intra-card context parallelism should be applied to each workload.

To accommodate sequences of different lengths, we parameterize CP partitioning by the maximum partition length $l_{\max}$, measured in chunks. Each sequence is divided into consecutive partitions of at most $l_{\max}$ chunks, with the final partition potentially shorter. All sequences in a batch share the same $l_{\max}$, while the number of partitions for each sequence depends on its length.

\subsection{Kernel Latency Modeling}
\label{sec:auto-cp-model}

We model the execution of each GDN kernel as a set of cooperative thread arrays (CTAs), each performing a workload-dependent number of sequential computation steps. Let $x$ denote an execution configuration, specifying the input shapes, whether CP is enabled, and the partition length when applicable. For kernel $\kappa$ under configuration $x$, let $k_{\kappa,r}(x)$ denote the number of steps executed by CTA $r$, and let $R_\kappa(x)$ denote the total number of CTAs.

Context parallelism affects these workloads in different ways. For recurrent and preparation kernels, using finer sequence partitions reduces the number of sequential steps while increasing the number of CTAs. In contrast, the sequential correction stage becomes longer as the number of partitions increases. Selecting an appropriate partition size therefore requires balancing these opposing effects.

Auto CP approximates kernel latency using the critical schedule depth, the total CTA count, and a fixed overhead term:
\begin{equation}
\widehat{t}_\kappa(x)
=
a_\kappa d_\kappa(x)
+
b_\kappa R_\kappa(x)
+
c_\kappa,
\label{eq:auto-cp-kernel-model}
\end{equation}
where $d_\kappa(x)$ denotes the critical schedule depth. The coefficients $a_\kappa$, $b_\kappa$, and $c_\kappa$ are fitted independently for each kernel family and hardware configuration.

Under a simplified scheduling model, we approximate the critical schedule depth by greedily assigning CTAs to the SM with the smallest accumulated step count and taking the maximum accumulated workload across SMs:
\begin{equation}
d_\kappa(x)
:=
\max_{s=1,\ldots,N_{\mathrm{sm}}}
\sum_{r\in\mathcal{A}_s(x)}
k_{\kappa,r}(x),
\label{eq:auto-cp-depth}
\end{equation}
where $N_{\mathrm{sm}}$ denotes the number of SMs and $\mathcal{A}_s(x)$ denotes the set of CTAs assigned to SM $s$.

Each execution configuration comprises a sequence of kernel invocations with dependencies between stages. Let $\mathcal{K}(x)$ denote the sequence of kernel invocations required by configuration $x$. Its total latency is estimated as
\begin{equation}
\widehat{T}(x)
=
\sum_{\kappa\in\mathcal{K}(x)}
\widehat{t}_\kappa(x).
\label{eq:auto-cp-total-model}
\end{equation}

The model coefficients are fitted offline using least squares over a small set of measured configurations. Once calibrated, the model provides inexpensive online latency estimates for candidate configurations.

\subsection{Wave-Aware Partition Search}
\label{sec:auto-cp-search}

For a given workload, Auto CP seeks the subsequence length $l_{\max}$ that minimizes $\widehat{T}(x(l_{\max}))$, where $x(l_{\max})$ denotes the CP configuration with partition length $l_{\max}$. Evaluating every supported length in $\mathcal{L}$, however, increases online search overhead. Since kernel latency can change sharply when the CTA count crosses execution-wave boundaries, Auto CP focuses on candidate lengths near these boundaries.

For a batch of $B$ sequences with lengths $L_b$ and $H$ independent heads, the recurrent kernel assigns one CTA to each partition-head pair. Using this layout, the subsequence length that approximately fills $w$ execution waves is
\begin{equation}
l_{\max}^{(w)}
\approx
\frac{H\sum_{b=1}^{B}L_b}
{wN_{\mathrm{sm}}}.
\label{eq:auto-cp-wave-solution}
\end{equation}
This estimate chooses a partition length that distributes the workload across approximately $wN_{\mathrm{sm}}$ CTAs, filling $w$ execution waves.

Auto CP enumerates wave counts $w$ in ascending order and maps each resulting $l_{\max}^{(w)}$ to nearby supported lengths in $\mathcal{L}$, stopping when the candidate set reaches a predefined size limit. The latency model then selects the best length from the resulting set $\mathcal{L}_{\mathrm{wave}}$:
\begin{equation}
l_{\max}^{*}
=
\underset{l_{\max}\in\mathcal{L}_{\mathrm{wave}}}
{\arg\min}
\;
\widehat{T}\!\left(x(l_{\max})\right).
\label{eq:auto-cp-best-candidate}
\end{equation}

This bounded search limits online decision overhead. Section~\ref{sec:auto-cp-evaluation} shows that the configurations selected by Auto CP achieve latency close to the exhaustive optimum, supporting a practical tradeoff between search cost and execution performance.

\subsection{CP Enablement}
\label{sec:auto-cp-decision}

When the workload already provides sufficient parallelism, CP offers limited benefit while introducing additional preprocessing and scheduling overhead. Auto CP therefore compares the best predicted CP configuration against the original no-CP execution and enables CP only when it is predicted to reduce latency.

Auto CP selects between the best CP configuration
$x(l_{\max}^{*})$ and the no-CP configuration $x_{\varnothing}$
according to their predicted latencies:
\begin{equation}
x^{*}
=
\begin{cases}
x(l_{\max}^{*}),
& \widehat{T}(x(l_{\max}^{*})) < \widehat{T}(x_{\varnothing}),\\
x_{\varnothing},
& \text{otherwise}.
\end{cases}
\label{eq:auto-cp-enable-rule}
\end{equation}

Auto CP thus jointly determines whether intra-card context parallelism should be enabled and, when beneficial, selects the partition length.

\section{Dual CP}
\label{sec:inter-intra-cp}

Scaling GDN to longer contexts motivates extending sequence parallelism beyond a single GPU. Existing inter-card methods such as KCP assign one sequence partition to each GPU, compute local state and transition summaries, and exchange them through an all-gather for boundary-state correction. However, inter-card CP alone may not provide sufficient parallelism within each GPU when the batch size and head count are small. Intra-card CP is therefore needed to expose additional sequence-level parallelism within each GPU.

A straightforward approach is to first perform inter-card CP to determine the input boundary state of each GPU, and then apply intra-card CP to its local sequence. However, the two levels independently perform state-summary preparation over the same chunks, duplicating the computation of state summaries $\widetilde{\mathbf{S}}$ and transition products $\mathbf{M}$.

\begin{figure}[t]
\centering
\includegraphics[width=\linewidth]{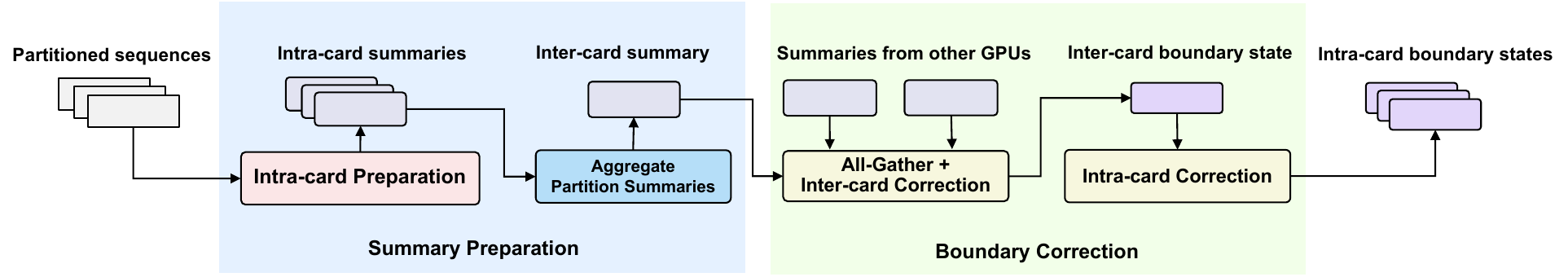}
\caption{
\textbf{Overview of Dual CP.}
Intra-card partition summaries are computed once and aggregated into inter-card summaries, enabling shared summary computation and hierarchical boundary-state correction without duplicated preparation.
}
\label{fig:dual-cp-overview}
\end{figure}

We therefore propose \emph{Dual CP}, which performs shared summary computation over intra-card partitions and derives inter-card summaries through lightweight aggregation. These summaries enable hierarchical boundary-state correction at both levels, avoiding duplicated preparation while exposing additional sequence-level parallelism within each GPU. Figure~\ref{fig:dual-cp-overview} illustrates the shared computation and hierarchical correction.

\begin{figure}[t]
\centering
\includegraphics[width=0.95\linewidth]{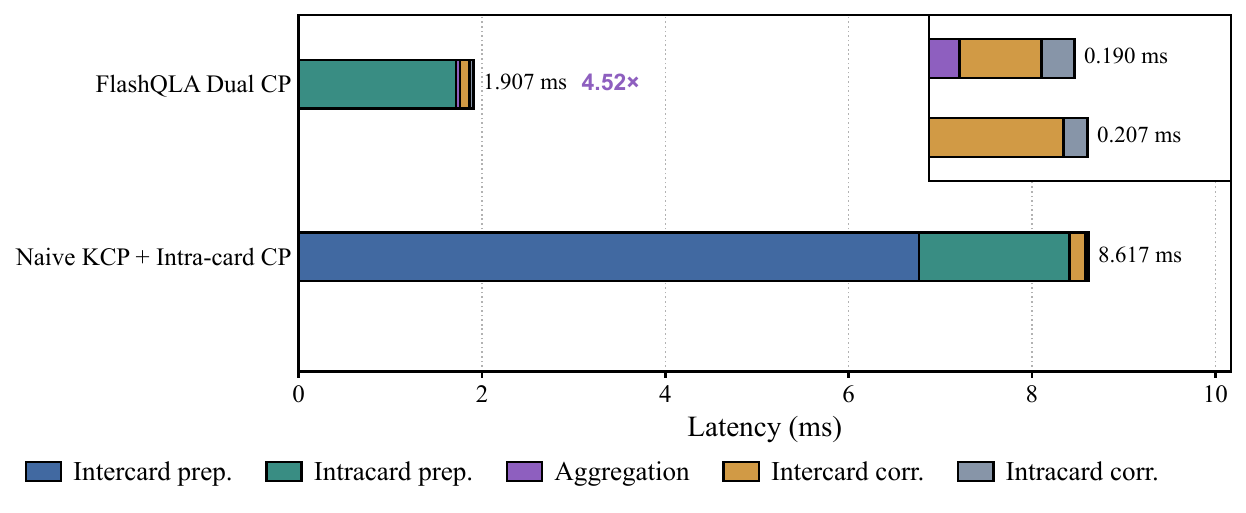}
\caption{
\textbf{Preprocessing latency breakdown of Dual CP.}
Comparison with a naive combination of inter-card and intra-card CP on eight GPUs with a 1M-token sequence and 32 attention heads.
Dual CP reduces total preprocessing latency from 8.617 ms to 1.907 ms ($4.52\times$).
The upper-right panel magnifies the smaller aggregation and correction stages.
}
\label{fig:dual-cp-preprocessing}
\end{figure}

Figure~\ref{fig:dual-cp-preprocessing} compares the preprocessing latency of Dual CP against the naive combination of inter-card and intra-card CP on eight GPUs with a 1M-token sequence and 32 attention heads. The naive approach incurs substantial overhead from duplicated state-summary preparation, whereas Dual CP eliminates this redundancy through shared summary computation, with only minor overhead from aggregation. As a result, Dual CP reduces total preprocessing latency from 8.617 ms to 1.907 ms, achieving a $4.52\times$ speedup.

\subsection{Shared Summary Computation}
\label{sec:dual-cp-preparation}

Let $N_{\mathrm{inter}}$ denote the number of GPUs. GPU $i$ owns an inter-card partition of $l_i$ chunks, divided into $N_{\mathrm{intra}}^i$ ordered intra-card partitions with lengths $l_{i,j}$, where
$l_i=\sum_{j=1}^{N_{\mathrm{intra}}^i}l_{i,j}$.
We use superscripts $(i)$ and $(i,j)$ to identify inter-card and intra-card partitions, respectively, while bracketed subscripts index chunks within each partition.
Following Section~\ref{sec:cp-theory}, each intra-card partition independently computes a zero-initialized terminal state
$\widetilde{\mathbf{S}}_{[l_{i,j}]}^{(i,j)}$
and a cumulative transition
$\mathbf{M}_{[l_{i,j}\leftarrow 1]}^{(i,j)}$.
Together, these summaries characterize the affine state transition:
\begin{equation}
\mathbf{S}_{[l_{i,j}]}^{(i,j)}
=
\widetilde{\mathbf{S}}_{[l_{i,j}]}^{(i,j)}
+
\mathbf{M}_{[l_{i,j}\leftarrow 1]}^{(i,j)}
\mathbf{S}_{[0]}^{(i,j)}.
\label{eq:dual-cp-local-affine}
\end{equation}

Since each local summary represents an affine state transition, the summaries can be composed to construct the inter-card summary without reprocessing the original chunks. Let $L_{i,j}:=\sum_{\ell=1}^{j}l_{i,\ell}$ denote the cumulative length of the first $j$ intra-card partitions on GPU $i$, with $L_{i,0}:=0$ and $L_{i,N_{\mathrm{intra}}^i}=l_i$. Initializing the empty-prefix summary with $\widetilde{\mathbf{S}}_{[0]}^{(i)}=\mathbf{0}$ and $\mathbf{M}_{[0\leftarrow 1]}^{(i)}=\mathbf{I}$, we aggregate the summaries in partition order:
\begin{align}
\widetilde{\mathbf{S}}_{[L_{i,j}]}^{(i)}
&=
\widetilde{\mathbf{S}}_{[l_{i,j}]}^{(i,j)}
+
\mathbf{M}_{[l_{i,j}\leftarrow 1]}^{(i,j)}
\widetilde{\mathbf{S}}_{[L_{i,j-1}]}^{(i)},
\label{eq:dual-cp-aggregate-s}
\\
\mathbf{M}_{[L_{i,j}\leftarrow 1]}^{(i)}
&=
\mathbf{M}_{[l_{i,j}\leftarrow 1]}^{(i,j)}
\mathbf{M}_{[L_{i,j-1}\leftarrow 1]}^{(i)}.
\label{eq:dual-cp-aggregate-m}
\end{align}
After aggregating all $N_{\mathrm{intra}}^i$ partitions, we obtain
$\widetilde{\mathbf{S}}_{[L_{i,N_{\mathrm{intra}}^i}]}^{(i)}
=\widetilde{\mathbf{S}}_{[l_i]}^{(i)}$
and
$\mathbf{M}_{[L_{i,N_{\mathrm{intra}}^i}\leftarrow 1]}^{(i)}
=\mathbf{M}_{[l_i\leftarrow 1]}^{(i)}$,
which form the state summaries of the entire inter-card partition.

Thus, Dual CP derives inter-card summaries by composing the intra-card affine summaries, eliminating the need for a separate preparation pass over the original chunks. Since aggregation operates on partition-level summaries rather than individual chunks, its cost scales with the number of intra-card partitions and introduces only limited overhead.

\subsection{Hierarchical Boundary-State Correction}
\label{sec:dual-cp-correction}

After aggregation, Dual CP performs hierarchical boundary-state correction. An all-gather exchanges the inter-card summaries. After all-gather, each GPU locally evaluates the ordered inter-card affine recurrence to obtain its own initial state. These states then initialize intra-card correction to compute the input states of local partitions.

For inter-card correction, the input states are computed from the gathered summaries using the affine recurrence, starting from
$\mathbf{S}_{[0]}^{(1)}=\mathbf{S}_0$:
\begin{equation}
\mathbf{S}_{[0]}^{(i+1)}
=
\widetilde{\mathbf{S}}_{[l_i]}^{(i)}
+
\mathbf{M}_{[l_i\leftarrow 1]}^{(i)}
\mathbf{S}_{[0]}^{(i)},
\qquad
i=1,\ldots,N_{\mathrm{inter}}-1.
\label{eq:dual-cp-inter-correction}
\end{equation}

For intra-card correction, each GPU initializes
$\mathbf{S}_{[0]}^{(i,1)}=\mathbf{S}_{[0]}^{(i)}$
and applies the same relation to its local summaries:
\begin{equation}
\mathbf{S}_{[0]}^{(i,j+1)}
=
\widetilde{\mathbf{S}}_{[l_{i,j}]}^{(i,j)}
+
\mathbf{M}_{[l_{i,j}\leftarrow 1]}^{(i,j)}
\mathbf{S}_{[0]}^{(i,j)},
\qquad
j=1,\ldots,N_{\mathrm{intra}}^i-1.
\label{eq:dual-cp-intra-correction}
\end{equation}
The local corrections proceed independently across GPUs. Once the initial states have been computed, all partitions can execute the main computation in parallel.

\section{Implementation and Kernel Design}
\label{sec:kernel-codesign}

FlashQLA provides an efficient GPU kernel implementation for context-parallel GDN by combining the execution characteristics of CP with kernel-level optimization techniques. The implementation integrates algebraic simplification, CP-aware kernel fusion, and manual warp specialization.

\subsection{Algebraic Simplification}
\label{sec:algebraic-simplification}

In the chunk-wise GDN formulation, cumulative gate factors appear both inside the triangular inverse in Eq.~\eqref{eq:gdn-chunk-a} and in the surrounding state-update operations. FlashQLA factors these terms out of the inverse to expose algebraic cancellations and eliminate redundant computation.

We define the gate-independent triangular inverse
\begin{equation}
\mathbf{A}_{[c]}^{\prime}
=
\left(
\mathbf{I}
+
\operatorname{StrictLower}
\left(
\mathbf{B}_{[c]}
\mathbf{K}_{[c]}\mathbf{K}_{[c]}^{\top}
\right)
\right)^{-1}.
\label{eq:gate-factored-aprime}
\end{equation}
The original gate-dependent inverse $\mathbf{A}$ can then be expressed as
\begin{equation}
\mathbf{A}
=
\mathbf{\Gamma}_{[c]}
\mathbf{A}_{[c]}^{\prime}
\mathbf{\Gamma}_{[c]}^{-1}.
\label{eq:gate-factored-a-relation}
\end{equation}

Substituting Equation~\eqref{eq:gate-factored-a-relation} into the chunk-wise state update eliminates redundant gate-dependent scaling and simplifies the computation of chunk transitions. Appendix~\ref{app:gate-reformulation} provides the derivation and the resulting state-update equations.

This reformulation also improves the numerical stability of cumulative transition computation during boundary-state preprocessing. Specifically, the chunk transition can be factorized as
\begin{equation}
\mathbf{M}_{[c]}
=
\gamma_{[c],C}
\underbrace{
\left(
\mathbf{I}
-
\mathbf{K}_{[c]}^\top
\mathbf{A}'_{[c]}
\mathbf{B}_{[c]}
\mathbf{K}_{[c]}
\right)
}_{\overline{\mathbf{M}}_{[c]}}.
\label{eq:gate-factored-transition}
\end{equation}
By separating the gate factors from the transition matrices, FlashQLA prevents cumulative gate decay from being repeatedly incorporated into low-precision matrix updates.

However, directly accumulating gate-independent transitions through low-precision matrix multiplications repeatedly rounds the cumulative result, introducing numerical errors over long partitions. FlashQLA instead reformulates transition accumulation as an additive update, allowing low-precision matrix multiplication while preserving the cumulative matrix in FP32:
\begin{equation}
\overline{\mathbf{M}}_{[j\leftarrow1]}^{(i)}
=
\overline{\mathbf{M}}_{[j-1\leftarrow1]}^{(i)}
-
\mathbf{K}_{[j]}^{(i)\top}
\mathbf{A}_{[j]}^{\prime(i)}
\mathbf{B}_{[j]}^{(i)}
\mathbf{K}_{[j]}^{(i)}
\overline{\mathbf{M}}_{[j-1\leftarrow1]}^{(i)}.
\label{eq:transition-accumulation}
\end{equation}
This formulation allows the cumulative transition to remain in FP32 accumulators across updates, avoiding repeated low-precision rounding while retaining efficient matrix multiplication. Appendix~\ref{app:transition-stability} provides the derivation and numerical stability analysis.

\subsection{CP-Aware Kernel Fusion}
\label{sec:cp-aware-fusion}

Context-parallel GDN involves computation stages with different parallelism and synchronization requirements. FlashQLA therefore organizes GPU kernels at chunk, CP-partition, and sequence granularities, fusing compatible operations while preserving synchronization between stages.

In the forward pass, chunk-local operations, including $\mathbf{K}\mathbf{K}^{\top}$ computation and the extended WY triangular solve, are fused into a single kernel using a chunk-level grid. Boundary-state preparation and correction are implemented as separate kernels because they operate at different parallelism granularities: preparation uses CP-partition-level grids, while correction uses sequence-level grids to resolve dependencies across partitions. Once boundary states are available, recurrent-state propagation and output computation are fused into a single kernel at the CP-partition level.

This stage-aware fusion reduces intermediate materialization and kernel launch overhead while aligning the execution granularity of each kernel with the parallelism exposed by CP. Batch and head dimensions provide additional parallelism across these stages.

\subsection{Manual Warp Specialization}
\label{sec:warp-specialization}

FlashQLA employs manual warp specialization in TileLang to overlap data movement, matrix multiplication, and element-wise computation within fused kernels. Dedicated producer warps preload data for subsequent computation, while consumer warp groups process the available data. The consumer schedule further overlaps matrix operations with independent element-wise computations where dependencies permit.

This design exploits intra-kernel concurrency and complements the sequence-level parallelism exposed by CP. Further implementation details, including pipeline schedules and warp-group assignments, are available in the FlashQLA technical blog~\cite{flashqla_blog}.

\section{Evaluation}
\label{sec:evaluation}

We evaluate FlashQLA across a range of workloads with varying sequence lengths, batch sizes, and head counts. We first compare its single- and multi-GPU performance against existing implementations, followed by an optimization breakdown to quantify the contributions of individual techniques. We then analyze Auto CP through latency-model accuracy, configuration-selection quality, and performance comparisons against static partitioning strategies.

\subsection{Experimental Setup}
\label{sec:experimental-setup}

All experiments are conducted on a server equipped with eight GPUs. We compare FlashQLA (v0.1.4), implemented with TileLang (v0.1.12)~\citep{wang2025tilelang}, against Flash Linear Attention (FLA, v0.5.1)~\citep{yang2024fla}. Unless otherwise specified, inputs use BFloat16 with $d_k=d_v=128$. Workload-specific batch sizes, head counts, and sequence lengths are reported in the corresponding experiments.

For synthetic benchmarks, we generate random inputs with heterogeneous gate-decay behavior across heads. Based on the prevalence of short decay windows observed in real models (Section~\ref{sec:gate-decay-observation}), we adopt a representative configuration in which 75\% of heads exhibit strong decay and the remaining 25\% exhibit weaker decay. Real-model experiments use activations and gate values collected from Qwen3.8-2.4T-A95B execution.

All reported latencies include the complete execution of the corresponding gated delta rule implementation, including boundary-state preprocessing and inter-GPU communication where applicable. We measure execution latency using CUDA events after warm-up and report the median across repeated runs.

\subsection{Single-Card Performance}
\label{sec:single-card-performance}

We compare FlashQLA with FLA for forward and backward execution using workloads derived from different Qwen model architectures and tensor-parallel (TP) configurations. For forward execution, we evaluate sequence lengths from 8K to 32K tokens under varying per-GPU head counts. For backward execution, we randomly generate batches of variable-length sequences with a fixed total of 32K tokens to examine performance under varying degrees of sequence-level concurrency.

\begin{figure}[t]
\centering
\includegraphics[width=0.95\linewidth]{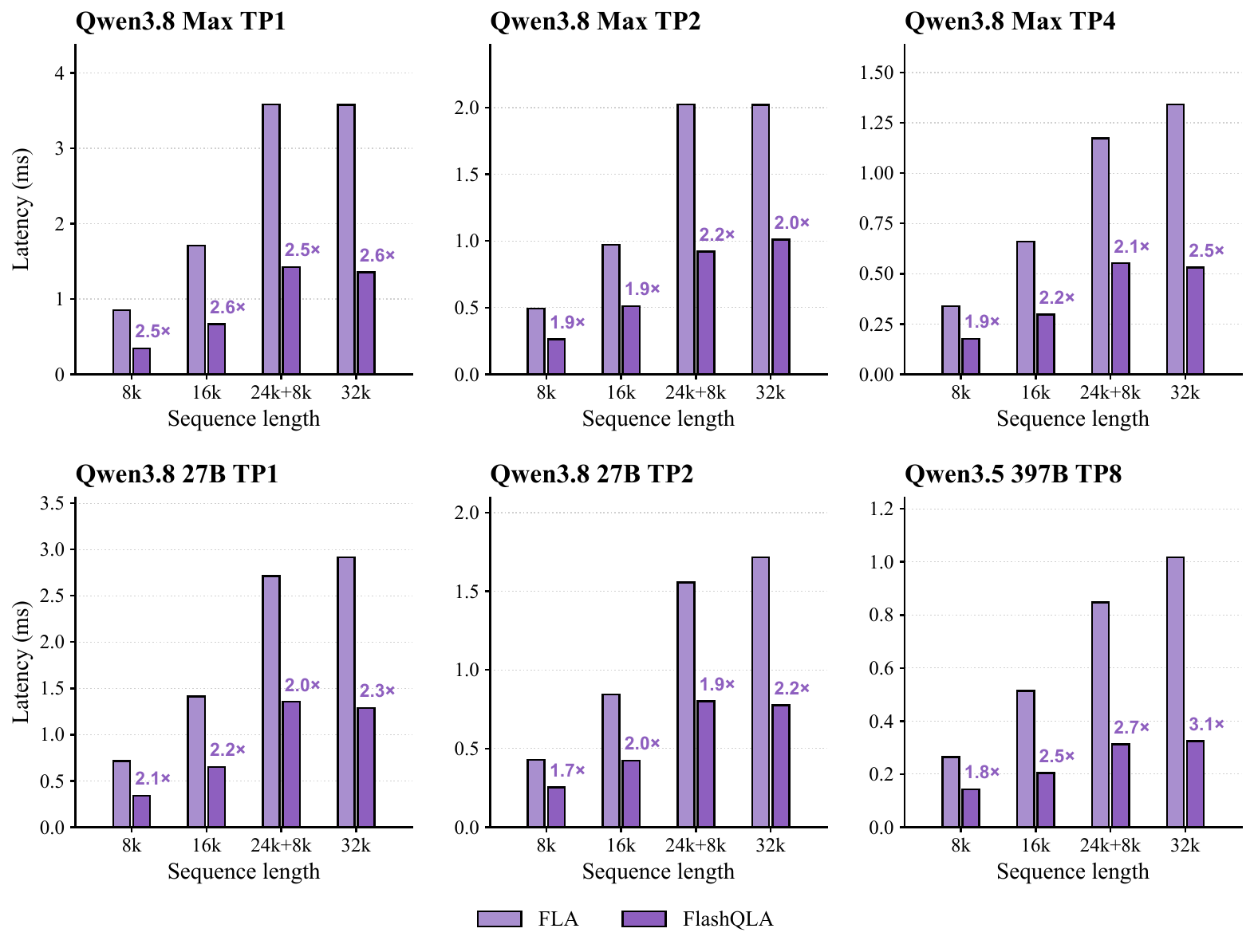}
\caption{
\textbf{Single-GPU forward performance.}
Comparison of FlashQLA and FLA forward performance across model architectures, tensor-parallel configurations, and sequence lengths from 8K to 32K tokens.
FlashQLA achieves $1.7\times$--$3.1\times$ speedups, with larger gains at higher TP degrees and longer sequence lengths.
}
\label{fig:single-card-forward}
\end{figure}

\begin{figure}[t]
\centering
\includegraphics[width=0.95\linewidth]{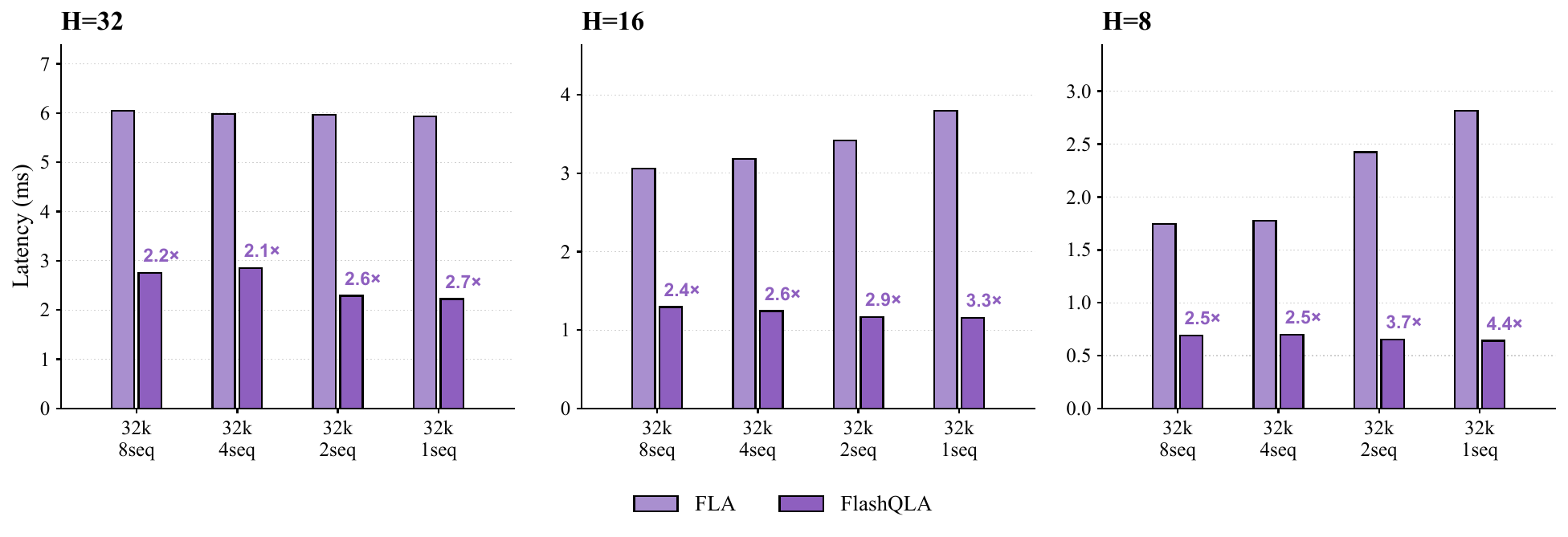}
\caption{
\textbf{Single-GPU backward performance.}
Comparison of FlashQLA and FLA backward performance across model architectures and tensor-parallel configurations, using a fixed total of 32K tokens randomly distributed among 1, 2, 4, or 8 sequences.
FlashQLA maintains stable latency across different sequence counts, while FLA becomes slower with fewer sequences. FlashQLA achieves over $2\times$ speedup in most configurations, reaching up to $4.4\times$.
}
\label{fig:single-card-backward}
\end{figure}

Figures~\ref{fig:single-card-forward} and~\ref{fig:single-card-backward} show that FlashQLA consistently outperforms FLA across the evaluated workloads, achieving forward speedups of $1.7\times$--$3.1\times$ and backward speedups above $2\times$ in most configurations, with a maximum of $4.4\times$. Forward gains are more pronounced at higher TP degrees and longer sequence lengths. Higher TP degrees reduce the per-GPU head count, making intra-card CP more beneficial, while longer sequences introduce longer inter-chunk dependency chains that can be shortened through sequence partitioning.

For backward execution, FlashQLA maintains relatively stable latency as the number of sequences varies under a fixed total token count of 32K. In contrast, FLA exhibits greater latency variation and becomes slower with fewer sequences, as reduced inter-sequence-level concurrency leaves more inter-chunk computation serialized. By exposing additional parallelism within each sequence, FlashQLA reduces its sensitivity to sequence count, which help mitigate pipeline bubbles in pipeline-parallel training.

To validate these gains on real model workloads, we further evaluate FlashQLA on Qwen3.8-2.4T-A95B using activations and gate values collected from actual model execution. We measure the average forward latency of the gated delta rule operator across all GDN layers at context lengths of 32K and 64K tokens under TP8, comparing FlashQLA against FLA with identical inputs and execution configurations.

\begin{table}[t]
    \centering
    \caption{
    \textbf{Real-model gated delta rule performance on Qwen3.8-2.4T-A95B.}
    Average forward latency of the gated delta rule operator across GDN layers using real model activations and gate values under TP8.
    }
    \label{tab:real-model-performance}
    \begin{tabular}{lrrr}
        \toprule
        \textbf{Seq. Length} &
        \textbf{FLA (ms)} &
        \textbf{FlashQLA (ms)} &
        \textbf{Speedup} \\
        \midrule
        32K & 2.309 & 0.711 & $3.25\times$ \\
        64K & 4.630 & 1.299 & $3.56\times$ \\
        \bottomrule
    \end{tabular}
\end{table}

As shown in Table~\ref{tab:real-model-performance}, FlashQLA achieves $3.25\times$ and $3.56\times$ forward speedups at 32K and 64K tokens, respectively. The gains are consistent with our synthetic benchmarks and increase with sequence length, demonstrating the effectiveness of FlashQLA on real model activations and gate distributions.

At the full-model level, FlashQLA reduces time to first token (TTFT) by approximately $5\%$ on Qwen3.8-2.4T-A95B with a 32K-token context under TP8. The smaller end-to-end gain is due to the relatively small fraction of TTFT spent on gated delta rule computation.

\subsection{Multi-Card Performance}
\label{sec:dual-cp-evaluation}

We evaluate multi-GPU execution on eight GPUs with total sequence lengths from 64K to 1M tokens and head counts $H\in\{16,32,64\}$. Each sequence is evenly partitioned across all eight GPUs. 

To distinguish the benefits of FlashQLA's execution optimizations, intra-card parallelism, and Dual CP, we compare four implementations:
\begin{enumerate}[leftmargin=*,itemsep=2pt,topsep=3pt]
    \item \textbf{FLA KCP}: the FLA implementation with inter-card CP (KCP).
    \item \textbf{FlashQLA KCP}: the FlashQLA kernel implementation with inter-card CP but without intra-card CP.
    \item \textbf{Naive KCP + Intra CP}: independent composition of inter-card and intra-card CP, with separate preprocessing at each level.
    \item \textbf{FlashQLA Dual CP}: joint inter-card and intra-card CP with shared summary computation and hierarchical correction.
\end{enumerate}

\begin{figure}[t]
\centering
\includegraphics[width=0.95\textwidth]{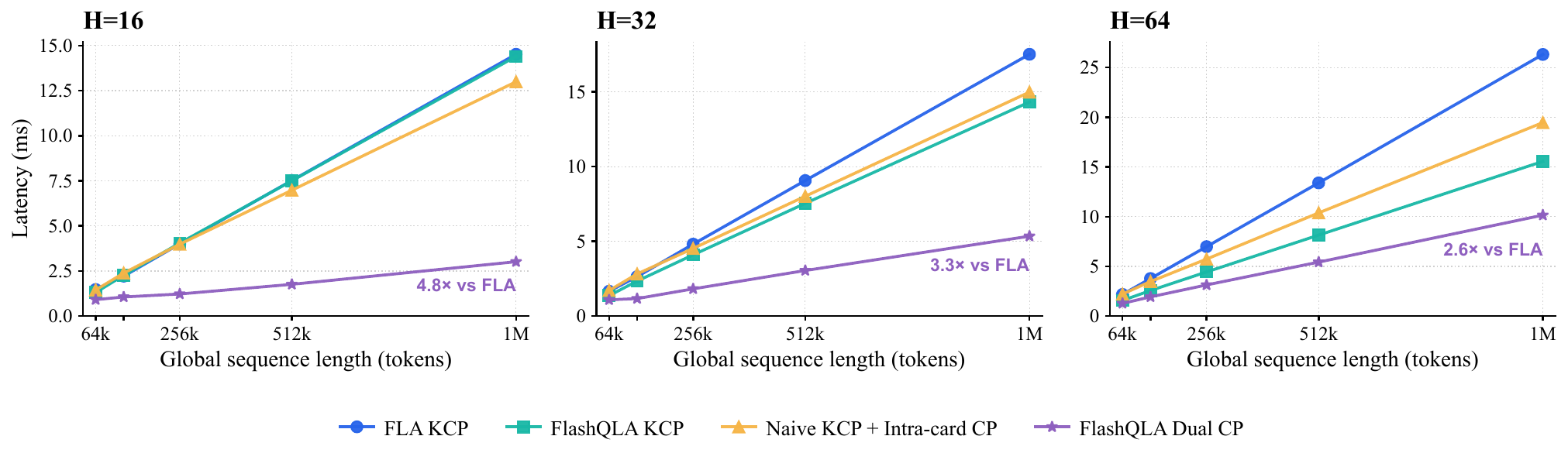}
\caption{
\textbf{Multi-GPU forward performance.}
Comparison of FLA KCP, FlashQLA KCP, naive KCP + intra-card CP, and FlashQLA Dual CP on eight GPUs across sequence lengths from 64K to 1M tokens and head counts of 16, 32, and 64.
Dual CP avoids the duplicated preprocessing of naive composition and achieves the largest gains at low head counts, where additional intra-card parallelism is most beneficial. FlashQLA Dual CP achieves up to $4.8\times$ speedup over FLA KCP.
}
\label{fig:dual-cp-forward}
\end{figure}

\begin{figure}[t]
\centering
\includegraphics[width=0.95\textwidth]{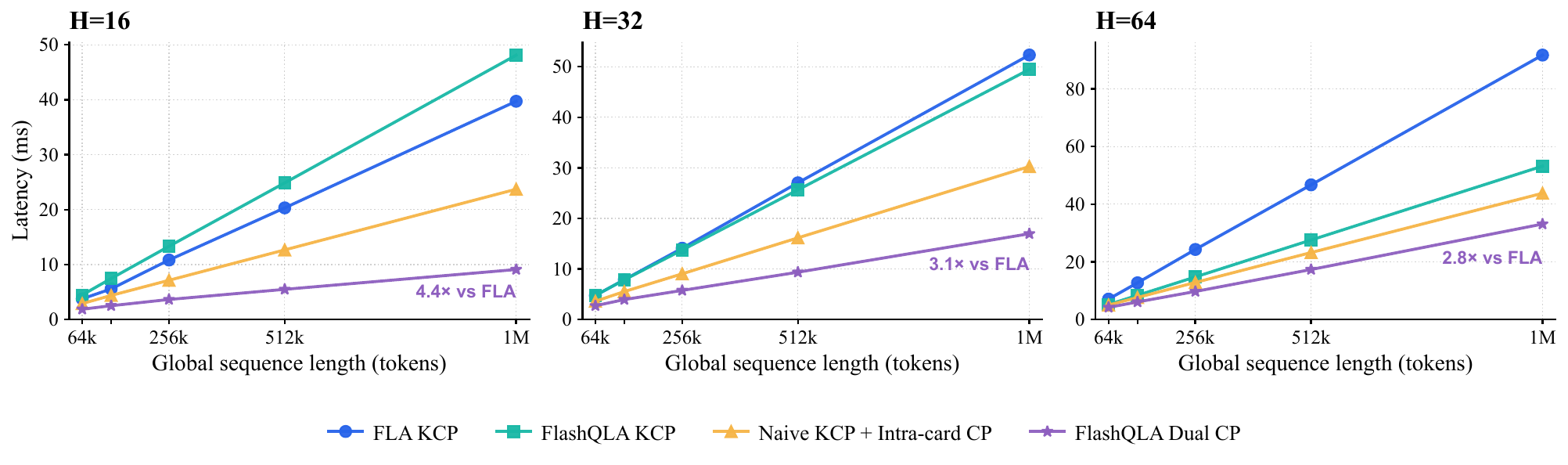}
\caption{
\textbf{Multi-GPU forward and backward performance.}
Combined forward-and-backward latency of four context-parallel implementations on eight GPUs.
Intra-card CP provides greater benefits than in forward-only execution, while Dual CP further reduces preprocessing overhead through shared summary computation. FlashQLA Dual CP achieves up to $4.4\times$ speedup over FLA KCP.
}
\label{fig:dual-cp-training}
\end{figure}

As shown in Figure~\ref{fig:dual-cp-forward}, FlashQLA KCP achieves approximately $1.5\times$--$2\times$ speedup over FLA KCP at $H=64$. At lower head counts, however, limited intra-card parallelism constrains the gains from kernel optimization, motivating the use of intra-card CP to expose additional sequence-level parallelism.

Introducing intra-card CP through Naive KCP + Intra CP provides additional sequence parallelism, but also duplicates preparation. At $H=16$, the increased parallelism improves performance, although the gain is constrained by the separate inter-card preparation, which still processes each GPU-local sequence sequentially. At $H=64$, the smaller gain from additional parallelism is outweighed by preprocessing overhead, causing the naive composition to underperform FlashQLA KCP.

Dual CP addresses this limitation by sharing preparation across the two levels. Deriving inter-card summaries from intra-card summaries avoids duplicated computation and exposes sequence parallelism during preparation. Relative to Naive KCP + Intra CP, Dual CP achieves up to $4.3\times$ speedup at $H=16$ and $2\times$ at $H=64$, demonstrating the importance of coordinating preparation when combining the two CP levels.

Figure~\ref{fig:dual-cp-training} reports combined forward-and-backward performance. The larger computational workload of the backward pass yields greater benefits from intra-card CP, allowing even Naive KCP + Intra CP to achieve more substantial gains than in forward-only execution. Dual CP further improves performance through shared summary computation and hierarchical correction, achieving approximately $1.2\times$--$2.6\times$ speedup over the naive composition across the evaluated configurations.

Overall, FlashQLA Dual CP achieves up to $4.8\times$ forward speedup and $4.4\times$ speedup for combined forward and backward execution over FLA KCP, demonstrating its effectiveness for long-context multi-GPU execution.

We further evaluate the strong-scaling performance of FLA KCP and FlashQLA Dual CP for combined forward and backward execution as the CP size increases from 1 to 8 GPUs. Each implementation is normalized to its own single-GPU latency.

\begin{figure}[t]
    \centering
    \includegraphics[width=1\textwidth]{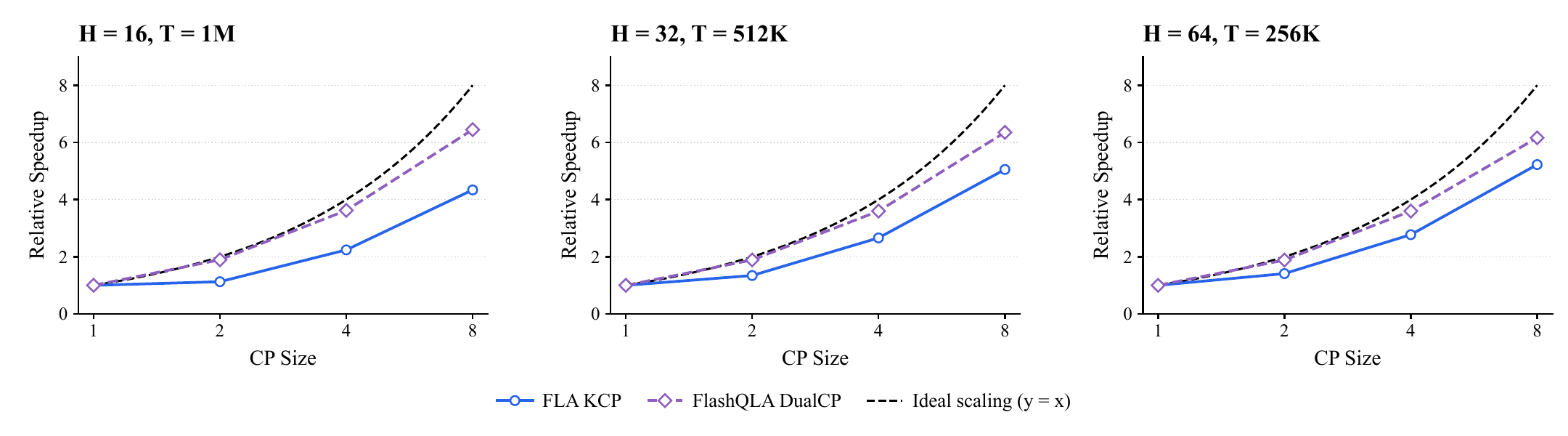}
    \caption{
    \textbf{Multi-GPU forward-and-backward scaling.}
    Relative speedup of FLA KCP and FlashQLA Dual CP across CP sizes of 1, 2, 4, and 8 GPUs, normalized to each implementation's single-GPU latency. The evaluated configurations are $H=16$ with 1M tokens, $H=32$ with 512K tokens, and $H=64$ with 256K tokens. The dashed line ($y=x$) indicates ideal linear scaling.
    }
    \label{fig:multi-gpu-scaling}
\end{figure}

As shown in Figure~\ref{fig:multi-gpu-scaling}, FlashQLA Dual CP achieves a $6.45\times$ speedup on eight GPUs with $H=16$, compared with $4.34\times$ for FLA KCP. FlashQLA Dual CP exhibits stronger scaling, approaching ideal linear scaling more closely by exposing additional intra-card parallelism and reducing redundant preparation overhead.

\subsection{Optimization Breakdown}
\label{sec:optimization-breakdown}

We examine the contributions of individual optimizations on a representative workload with $H=16$ and a sequence length of 1M tokens. Starting from the optimized FlashQLA kernels without CP, we incrementally enable intra-card CP with Auto CP and decay-aware preparation, and extend to Dual CP with 8 GPUs.

\begin{figure}[t]
\centering

\includegraphics[width=0.6\textwidth]{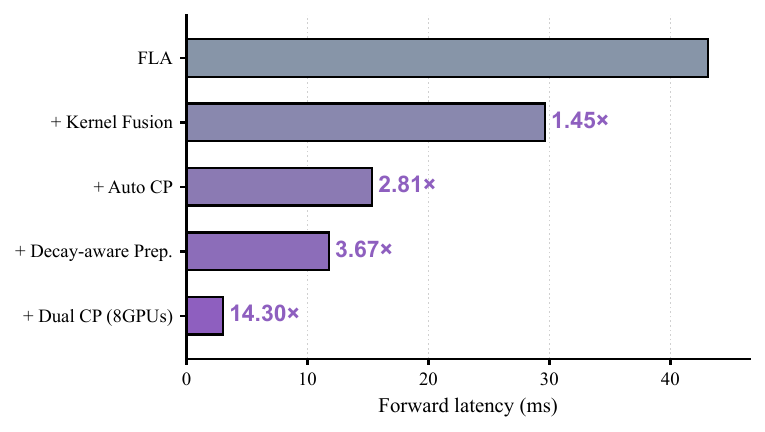}
\caption{
\textbf{Optimization breakdown of FlashQLA.}
Cumulative forward speedup relative to single-GPU FLA on GPUs with 16 heads and a 1M-token sequence.
The stages progressively incorporate optimized kernels, intra-card CP with Auto CP, decay-aware preparation, and eight-GPU Dual CP.
}
\label{fig:optimization-breakdown}
\end{figure}

As shown in Figure~\ref{fig:optimization-breakdown}, kernel optimization alone achieves a $1.45\times$ speedup over FLA. Enabling intra-card CP with Auto CP exposes additional sequence parallelism and increases the speedup to $2.81\times$, although the realized gains remain limited by boundary-state preprocessing overhead. Decay-aware preparation substantially reduces this overhead, further improving the speedup to $3.67\times$. Finally, scaling to eight GPUs with Dual CP achieves a $14.30\times$ speedup, combining multi-GPU parallelism with shared summary computation and hierarchical boundary-state correction.

\subsection{Auto CP Analysis}
\label{sec:auto-cp-evaluation}

Auto CP relies on latency predictions to select among a bounded set of partitioning configurations. We evaluate both the model's ability to capture partition-dependent performance trends and the measured performance of the resulting selections.

\paragraph{Latency-model accuracy.}
We evaluate the predictive accuracy of the kernel-level latency model introduced in Section~\ref{sec:auto-cp-model} using a variable-length workload with sequence lengths $[512, 2048, 32768]$ and $H=32$. We consider four core kernels: \texttt{prepare\_h}, \texttt{correct\_h}, \texttt{fused\_fwd}, and \texttt{fused\_bwd}. For this evaluation, each kernel model is calibrated using four measured configurations, with the remaining candidate partition lengths used to assess prediction accuracy.

\begin{figure}[t]
\centering
\includegraphics[width=0.8\textwidth]{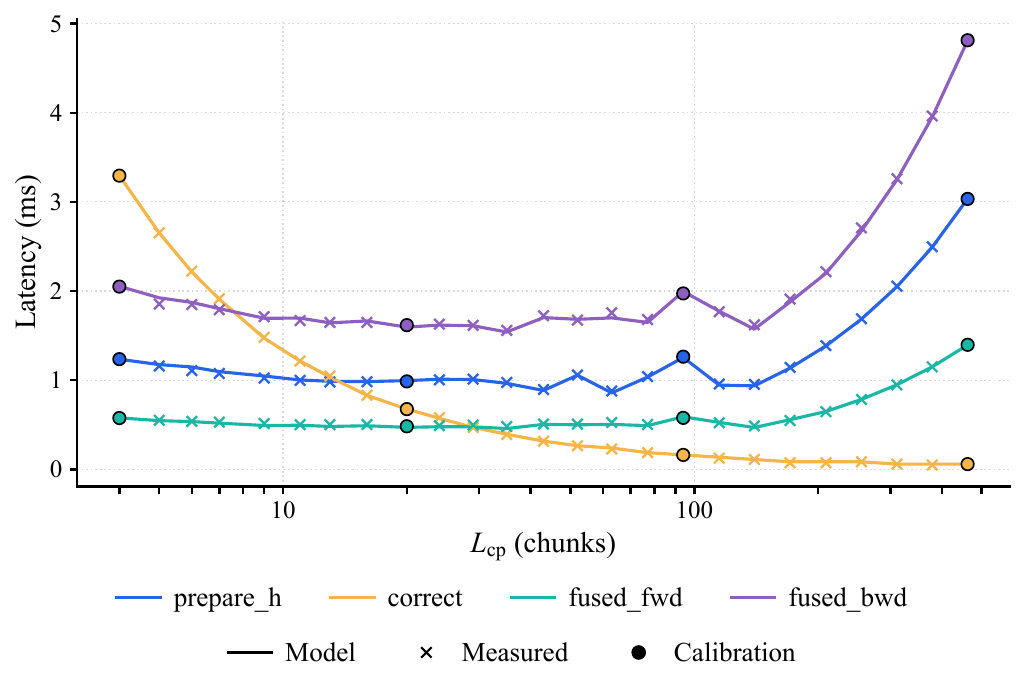}
\caption{
\textbf{Kernel latency-model accuracy.}
Model-predicted latency curves and measured latencies across candidate partition lengths for a variable-length workload with sequence lengths $[512, 2048, 32768]$ and 32 heads, covering all four kernels.
}
\label{fig:auto-cp-model-accuracy}
\end{figure}

As shown in Figure~\ref{fig:auto-cp-model-accuracy}, the measured latencies closely follow the model-predicted curves across all four kernels and a wide range of partition lengths. For \texttt{prepare\_h}, \texttt{fused\_fwd}, and \texttt{fused\_bwd}, the model captures latency fluctuations induced by execution-wave transitions, as well as the increased overhead of finer partitioning. For \texttt{correct\_h}, the model captures the increasing correction overhead as smaller partition lengths lead to more partitions. These results demonstrate that the latency model captures the key performance characteristics of different partitioning configurations, providing a reliable basis for Auto CP selection.

\paragraph{Configuration-selection quality.}
We next assess whether these predictions lead to efficient execution configurations. We evaluate 90 workloads spanning different sequence lengths, batch sizes, variable-length distributions, and head counts, under both forward and combined forward-and-backward execution.

For each workload, we exhaustively measure all supported CP configurations and the no-CP configuration to obtain the oracle configuration $x_{\mathrm{oracle}}^*$. Let $x^*$ denote the configuration selected by Auto CP and $T(x)$ the measured execution latency under configuration $x$. We quantify selection quality through relative latency regret:
\begin{equation}
\operatorname{Regret}
=
\frac{
T(x^*) - T(x_{\mathrm{oracle}}^*)
}{
T(x_{\mathrm{oracle}}^*)
}.
\label{eq:auto-cp-regret-eval}
\end{equation}
This metric measures the execution cost of the selected configuration rather than requiring an exact match with the oracle.

\begin{figure}[t]
\centering
\includegraphics[width=0.7\textwidth]{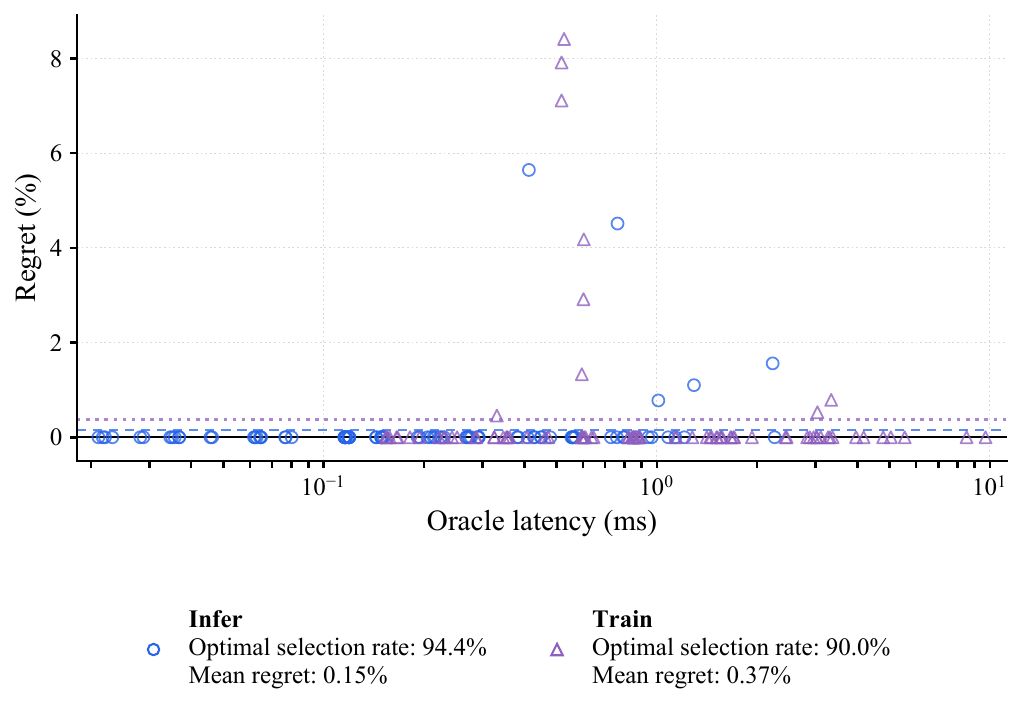}
\caption{
\textbf{Auto CP configuration-selection quality.}
Relative latency regret of Auto CP-selected configurations against the exhaustive-search oracle, plotted against workload latency across 90 workloads under inference (forward-only) and training (forward-and-backward) execution.
}
\label{fig:auto-cp-decision-quality}
\end{figure}

As shown in Figure~\ref{fig:auto-cp-decision-quality}, Auto CP selects the optimal configuration for $94.4\%$ of inference workloads and $90.0\%$ of training workloads, with mean latency regrets of only $0.15\%$ and $0.37\%$, respectively. These results demonstrate that latency modeling combined with bounded wave-aware search achieves near-optimal performance on average with a small candidate set.

\paragraph{Performance Benefit of Auto CP.}
We further evaluate the performance benefit of Auto CP against two static partitioning strategies: a fixed maximum partition length of $l_{\max}=32$ and a hardware-aware heuristic that sets $l_{\max}=\lceil N_{\mathrm{sm}}/(BH)\rceil$, where $N_{\mathrm{sm}}$ denotes the number of SMs, $B$ the batch size, and $H$ the number of attention heads. We measure the complete linear attention execution under both forward-only (inference) and combined forward-and-backward (training) settings across different sequence lengths from 1k to 32k and head counts. For each head count, we normalize the latency of each strategy to Auto CP and report the geometric mean across sequence lengths.

\begin{figure}[t]
\centering
\includegraphics[width=0.9\textwidth]{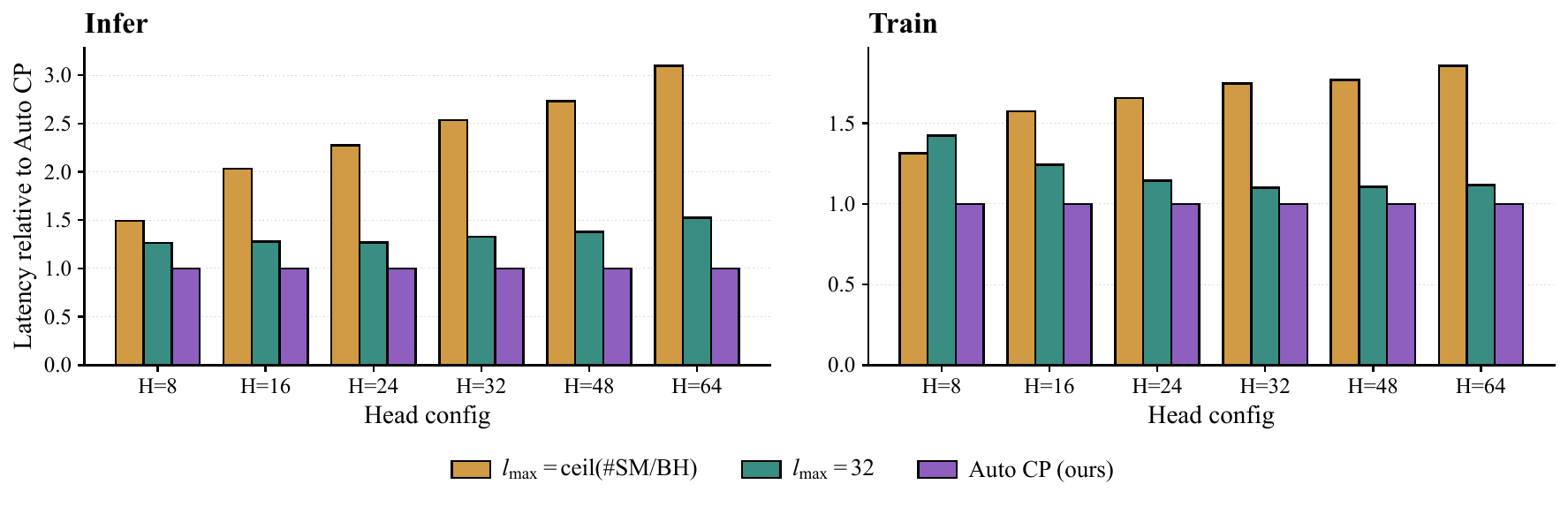}
\caption{
\textbf{Performance benefit of Auto CP over static partitioning.}
Geometric-mean relative latency of fixed partitioning and a hardware-aware heuristic across sequence lengths for each head count, normalized to Auto CP.
The left and right panels show forward-only and combined forward-and-backward execution, respectively.
Auto CP achieves the lowest latency across all evaluated settings, with larger performance gains under inference.
}
\label{fig:auto-cp-speedup}
\end{figure}

Figure~\ref{fig:auto-cp-speedup} shows that Auto CP consistently achieves the lowest latency among the evaluated strategies across all head counts and execution settings. Compared with fixed partitioning, Auto CP provides consistent improvements by adapting partition lengths to workload characteristics. The gains are more pronounced over the hardware-aware heuristic, reaching up to $3\times$ under inference, as the heuristic considers hardware parallelism but does not account for kernel efficiency and boundary-state preprocessing overhead. The gains are less pronounced in training, as the higher computational cost of the backward pass reduces the proportion of execution time spent on preprocessing. Overall, these results demonstrate the effectiveness of latency-aware partition selection across diverse workloads.
\section{Related Work}
\label{sec:related}

\paragraph{Recurrent Linear Attention.}
Linear attention replaces quadratic token-wise interactions with a recurrent fixed-size state, enabling linear-time sequence processing~\citep{katharopoulos2020transformers}. DeltaNet augments this formulation with a delta-rule update that selectively modifies the recurrent state and introduces an extended WY representation to parallelize computation within chunks~\citep{yang2024parallelizing}. Gated DeltaNet (GDN) further incorporates data-dependent state decay and develops an efficient chunk-wise GPU formulation~\citep{yang2025gated}. Kimi Delta Attention (KDA) extends GDN with fine-grained, channel-wise decay gates~\citep{zhang2025kda}. FlashQLA focuses on GDN, while its affine-summary-based context-parallel framework provides a foundation for extending sequence parallelism to other recurrent linear-attention architectures.

\paragraph{Efficient Linear Attention Execution and Context Parallelism.}
Flash Linear Attention (FLA) provides hardware-efficient Triton implementations for a broad range of linear-attention architectures and serves as our single-GPU baseline~\citep{yang2024fla}.
For distributed execution, Kimi K3 adopts KDA Context Parallelism (KCP), which partitions sequences across devices, computes local state and transition summaries, and determines partition initial states through boundary-state correction~\citep{kimi2026k3}.
FlashQLA complements these efforts by combining intra-card and inter-card sequence parallelism with shared summary computation and gate-driven preparation.

Following the open-source release of FlashQLA, optimized GDN implementations in FlashInfer~\citep{ye2025flashinfer} and NVIDIA cuDNN have incorporated similar algebraic reformulations and context-parallel strategies. These developments further demonstrate the practical relevance of combining algebraic simplification with sequence-level parallelism for efficient GDN execution.

\section{Conclusion}
\label{sec:conclusion}

We presented FlashQLA, a context-parallel execution framework for Gated DeltaNet that exposes sequence-level parallelism within and across GPUs while reducing boundary-state preprocessing overhead. FlashQLA exploits head-wise gate decay through decay-aware preparation and integrates intra- and inter-card parallelism through Dual CP with shared summary computation. Workload-adaptive Auto CP and CP-aware kernel co-design further improve execution efficiency.

FlashQLA achieves up to $3.1\times$ forward and $4.4\times$ backward speedup over FLA on a single GPU. Scaling to eight GPUs with contexts up to 1M tokens, Dual CP achieves up to $4.8\times$ forward speedup and $4.4\times$ speedup for combined forward and backward execution over FLA KCP. These results demonstrate that effective context parallelism requires not only exposing additional sequence parallelism but also reducing the preprocessing overhead needed to enable it.

\section*{Acknowledgments}

We would like to express our sincere gratitude to Hongjie Xie and other community developers for their contributions to the development and maintenance of FlashQLA kernels for advanced computing architectures. Their efforts have helped extend FlashQLA’s compatibility with and support across GPU architectures. We would also like to thank Guangyun Han and the Flash-Linear Attention and TileLang communities; the technical exchanges and discussions with them have been a great source of inspiration to us. Finally, we thank all contributors to the FlashQLA open-source project for their valuable contributions to code maintenance, engineering improvements, and the project’s continued development.

\bibliographystyle{plainnat}
\bibliography{flashqla}

\appendix

\section{Notation Summary}
\label{app:notation}

This appendix summarizes the notation used throughout the paper.
Ordinary subscripts denote token indices, while bracketed subscripts
denote chunk indices or boundaries. Superscripts identify partitions:
$(i)$ for a single-level partition or inter-card partition, and
$(i,j)$ for an intra-card partition on GPU $i$.

\begin{table*}[h]
\centering
\caption{Notation for Gated DeltaNet context parallelism and boundary-state preprocessing.}
\label{tab:notation-gdn}
\small
\setlength{\tabcolsep}{5pt}
\renewcommand{\arraystretch}{1.18}
\begin{tabular}{@{}ll@{\hspace{1.2cm}}ll@{}}
\toprule
\textbf{Symbol} & \textbf{Description} &
\textbf{Symbol} & \textbf{Description} \\
\midrule
\multicolumn{4}{l}{\textit{General notation}} \\
$C$ & Tokens per chunk &
$L$ & Chunks per sequence \\
$B$ & Batch size &
$L_b$ & Chunks in sequence $b$ \\
$H$ & Number of heads &
$d_k,d_v$ & Key/value dimensions \\
$t$ & Token index &
$c$ & Global chunk index \\
$h$ & Head index &
$\mathbf{S}_0$ & Initial recurrent state \\
\midrule
\multicolumn{4}{l}{\textit{Gated DeltaNet}} \\
$\mathbf{S}_t$ & State after token $t$ &
$\mathbf{o}_t$ & Output at token $t$ \\
$\mathbf{q}_t,\mathbf{k}_t,\mathbf{v}_t$ & Token-level Q/K/V &
$\alpha_t,\beta_t$ & Decay gate / update strength \\
$\mathbf{Q}_{[c]},\mathbf{K}_{[c]},\mathbf{V}_{[c]}$ & Chunk-level Q/K/V &
$\mathbf{O}_{[c]}$ & Chunk output \\
$\alpha_{[c],t},\beta_{[c],t}$ & Local token gates &
$\gamma_{[c],t}$ & Cumulative token decay \\
$\mathbf{B}_{[c]}$ & Diagonal update strengths &
$\boldsymbol{\Gamma}_{[c]}$ & Diagonal cumulative decay \\
$\mathbf{A}_{[c]}$ & Extended WY inverse &
$\mathbf{U}_{[c]},\mathbf{W}_{[c]}$ & Extended WY auxiliaries \\
$\mathbf{S}_{[c]}$ & State entering chunk $c$ &
$\mathbf{M}_{[c]}$ & Chunk transition matrix \\
$\mathbf{E}_{[c]}$ & Chunk-local contribution &
$\gamma_{[c],C}$ & Total chunk decay \\
\midrule
\multicolumn{4}{l}{\textit{Boundary-state preprocessing}} \\
$N$ & Number of partitions &
$l_i$ & Chunks in partition $i$ \\
$\mathbf{S}_{[j]}^{(i)}$ & State at local boundary $j$ &
$\mathbf{S}_{[0]}^{(i)}$ & Partition input state \\
$\mathbf{M}_{[j]}^{(i)}$ & Local chunk transition &
$\mathbf{E}_{[j]}^{(i)}$ & Local chunk contribution \\
$\widetilde{\mathbf{S}}_{[j]}^{(i)}$ & Zero-state rollout &
$\Delta\mathbf{S}_{[j]}^{(i)}$ & State difference due to the initial state \\
$\mathbf{M}_{[j\leftarrow 1]}^{(i)}$ & Cumulative transition &
$\mathbf{S}_{[l_i]}^{(i)}$ & Partition terminal state \\
\bottomrule
\end{tabular}
\end{table*}

\begin{table*}[h]
\centering
\caption{Notation for decay-aware preparation and Auto CP.}
\label{tab:notation-optimization}
\small
\setlength{\tabcolsep}{5pt}
\renewcommand{\arraystretch}{1.18}
\begin{tabular}{@{}ll@{\hspace{1.2cm}}ll@{}}
\toprule
\textbf{Symbol} & \textbf{Description} &
\textbf{Symbol} & \textbf{Description} \\
\midrule
\multicolumn{4}{l}{\textit{Decay-aware preparation}} \\
$\gamma_{[c],C,h}$ & Chunk decay for head $h$ &
$D_h(a,b)$ & Cumulative decay over $[a,b]$ \\
$\mathrm{TOL}$ & Decay threshold &
$K_{c,h}$ & Decay window in chunks \\
$W_{c,h}$ & Decay window in tokens &
$K_{\max}$ & Maximum truncated window \\
$K_i$ & Boundary decay window &
$\overline{\mathbf{M}}_{[c]}$ & Decay-normalized transition \\
$\Delta\mathbf{S}_{[c]}$ & State difference due to truncated history &
$\widetilde{\mathbf{S}}_{[l_{i-1}],K_i}^{(i-1)}$ &
Truncated zero-state rollout \\
\midrule
\multicolumn{4}{l}{\textit{Auto CP: latency proxy model}} \\
$x$ & Execution configuration &
$\kappa$ & Kernel identifier \\
$r$ & CTA index &
$R_\kappa(x)$ & Number of CTAs \\
$k_{\kappa,r}(x)$ & CTA computation steps &
$N_{\mathrm{sm}}$ & Number of SMs \\
$\mathcal{A}_s(x)$ & CTAs assigned to SM $s$ &
$d_\kappa(x)$ & Critical schedule depth \\
$a_\kappa,b_\kappa,c_\kappa$ & Fitted latency coefficients &
$\widehat{t}_\kappa(x)$ & Predicted kernel latency \\
$\mathcal{K}(x)$ & Kernel invocation sequence &
$\widehat{T}(x)$ & Predicted total latency \\
\midrule
\multicolumn{4}{l}{\textit{Auto CP: partition search and enablement}} \\
$l_{\max}$ & Maximum partition length &
$\mathcal{L}$ & Supported partition lengths \\
$w$ & Execution-wave count &
$l_{\max}^{(w)}$ & Wave-boundary length estimate \\
$\mathcal{L}_{\mathrm{wave}}$ & Wave-aware candidates &
$l_{\max}^{*}$ & Selected partition length \\
$x(l_{\max})$ & CP configuration &
$x_{\varnothing}$ & No-CP configuration \\
$x^{*}$ & Selected configuration &
& \\
\bottomrule
\end{tabular}
\end{table*}

\begin{table*}[h]
\centering
\caption{Notation for Dual CP.}
\label{tab:notation-dual-cp}
\small
\setlength{\tabcolsep}{5pt}
\renewcommand{\arraystretch}{1.18}
\begin{tabular}{@{}ll@{\hspace{1.2cm}}ll@{}}
\toprule
\textbf{Symbol} & \textbf{Description} &
\textbf{Symbol} & \textbf{Description} \\
\midrule
\multicolumn{4}{l}{\textit{Partition hierarchy}} \\
$N_{\mathrm{inter}}$ & Number of GPUs &
$N_{\mathrm{intra}}^i$ & Local partitions on GPU $i$ \\
$l_i$ & Chunks assigned to GPU $i$ &
$l_{i,j}$ & Chunks in local partition $(i,j)$ \\
$\mathbf{S}_{[c]}^{(i)}$ & GPU-local state &
$\mathbf{S}_{[c]}^{(i,j)}$ & Intra-card partition state \\
\midrule
\multicolumn{4}{l}{\textit{Shared Summary Computation}} \\
$\widetilde{\mathbf{S}}_{[l_{i,j}]}^{(i,j)}$ & Local zero-state terminal state &
$\mathbf{M}_{[l_{i,j}\leftarrow 1]}^{(i,j)}$ & Local cumulative transition \\
$L_{i,j}$ & Cumulative local chunk count &
$\widetilde{\mathbf{S}}_{[L_{i,j}]}^{(i)}$ & Aggregated zero-state summary \\
$\mathbf{M}_{[L_{i,j}\leftarrow 1]}^{(i)}$ & Aggregated transition &
$\widetilde{\mathbf{S}}_{[l_i]}^{(i)}$ & GPU-level zero-state summary \\
$\mathbf{M}_{[l_i\leftarrow 1]}^{(i)}$ & GPU-level transition &
& \\
\midrule
\multicolumn{4}{l}{\textit{Hierarchical boundary-state correction}} \\
$\mathbf{S}_{[0]}^{(i)}$ & GPU input boundary state &
$\mathbf{S}_{[0]}^{(i,j)}$ & Local partition input state \\
\bottomrule
\end{tabular}
\end{table*}

\section{Derivation of the Gate-Factored GDN Reformulation}
\label{app:gate-reformulation}

We derive the gate-factored GDN reformulation introduced in Section~\ref{sec:algebraic-simplification}, using the chunk-wise formulation defined in Section~\ref{sec:gdn-review}. The derivation shows how gate-dependent scaling can be factored out of the triangular inverse to simplify the chunk-wise state transition.

For a chunk $c$, recall from Equation~\eqref{eq:gdn-chunk-a} that
\begin{equation}
\mathbf{A}_{[c]}^{-1}
=
\mathbf{I}
+
\operatorname{StrictLower}\left(
\mathbf{B}_{[c]}
\boldsymbol{\Gamma}_{[c]}
\mathbf{K}_{[c]}\mathbf{K}_{[c]}^\top
\boldsymbol{\Gamma}_{[c]}^{-1}
\right).
\end{equation}
Since $\mathbf{B}_{[c]}$ and $\boldsymbol{\Gamma}_{[c]}$ are diagonal and diagonal scaling preserves the strictly lower-triangular structure, we have
\begin{align}
\mathbf{A}_{[c]}^{-1}
&=
\boldsymbol{\Gamma}_{[c]}
\left(
\mathbf{I}
+
\operatorname{StrictLower}\left(
\mathbf{B}_{[c]}
\mathbf{K}_{[c]}\mathbf{K}_{[c]}^\top
\right)
\right)
\boldsymbol{\Gamma}_{[c]}^{-1}
\nonumber\\
&=
\boldsymbol{\Gamma}_{[c]}
\mathbf{A}_{[c]}^{\prime-1}
\boldsymbol{\Gamma}_{[c]}^{-1}.
\label{eq:appendix-a-factorization}
\end{align}
Here, $\mathbf{A}_{[c]}^{\prime}$ is the gate-independent triangular inverse defined in Equation~\eqref{eq:gate-factored-aprime}. Inverting both sides establishes
\begin{equation}
\mathbf{A}_{[c]}
=
\boldsymbol{\Gamma}_{[c]}
\mathbf{A}_{[c]}^{\prime}
\boldsymbol{\Gamma}_{[c]}^{-1},
\label{eq:appendix-a-relation}
\end{equation}
which is Equation~\eqref{eq:gate-factored-a-relation}.

Substituting this factorization into the auxiliary quantities in Equations~\eqref{eq:gdn-chunk-u} and~\eqref{eq:gdn-chunk-w} gives
\begin{align}
\boldsymbol{\Gamma}_{[c]}^{-1}\mathbf{U}_{[c]}
&=
\mathbf{A}_{[c]}^{\prime}
\mathbf{B}_{[c]}
\boldsymbol{\Gamma}_{[c]}^{-1}
\mathbf{V}_{[c]},
\label{eq:appendix-u-simplified}\\
\boldsymbol{\Gamma}_{[c]}^{-1}\mathbf{W}_{[c]}
&=
\mathbf{A}_{[c]}^{\prime}
\mathbf{B}_{[c]}
\mathbf{K}_{[c]}.
\label{eq:appendix-w-simplified}
\end{align}
The second equality follows from the commutativity of $\mathbf{B}_{[c]}$ and ${\boldsymbol{\Gamma}}_{[c]}$. These expressions eliminate the gate-dependent scaling surrounding the triangular inverse.

Substituting Equations~\eqref{eq:appendix-u-simplified} and~\eqref{eq:appendix-w-simplified} into the chunk-level transition in Equation~\eqref{eq:chunk-summary}, we obtain
\begin{align}
\mathbf{M}_{[c]}
&=
\gamma_{[c],C}
\left(
\mathbf{I}
-
\mathbf{K}_{[c]}^\top
\mathbf{A}_{[c]}^{\prime}
\mathbf{B}_{[c]}
\mathbf{K}_{[c]}
\right),
\label{eq:appendix-m-simplified}\\
\mathbf{E}_{[c]}
&=
\gamma_{[c],C}
\mathbf{K}_{[c]}^\top
\mathbf{A}_{[c]}^{\prime}
\mathbf{B}_{[c]}
\boldsymbol{\Gamma}_{[c]}^{-1}
\mathbf{V}_{[c]}.
\label{eq:appendix-e-simplified}
\end{align}
Therefore, the chunk-wise state update retains its affine form
\begin{equation}
\mathbf{S}_{[c+1]}
=
\mathbf{M}_{[c]}\mathbf{S}_{[c]}
+
\mathbf{E}_{[c]},
\end{equation}
but its transition matrices can now be computed using the gate-independent triangular inverse $\mathbf{A}_{[c]}^{\prime}$.

Finally, the cumulative gate factors in $\mathbf{E}_{[c]}$ can be expressed as relative decay factors. For each local token position $t$,
\begin{equation}
\begin{aligned}
\frac{\gamma_{[c],C}}{\gamma_{[c],t}}
&=
\exp\left(
\log\gamma_{[c],C}
-
\log\gamma_{[c],t}
\right)\\
&=
\exp\left(
\sum_{j=t+1}^{C}\log\alpha_{[c],j}
\right),
\end{aligned}
\label{eq:appendix-relative-decay}
\end{equation}
where the empty sum equals zero. By computing the difference of cumulative log gates before exponentiation, we avoid explicitly forming potentially underflowed cumulative products and performing numerically unstable division.

Thus, the reformulation removes gate-dependent scaling from the triangular inverse while preserving the original chunk-wise state transition.

\section{Numerical Stability of Transition Accumulation}
\label{app:transition-stability}

During boundary-state preprocessing, FlashQLA computes the cumulative transition matrix over each partition. Since this involves repeated matrix multiplications across chunks, numerical errors may accumulate, particularly when low-precision matrix multiplications  are used. We improve numerical stability through gate separation and FP32-accumulated transition updates.

\subsection{Gate Separation}

Recall that the chunk transition can be factorized as
\begin{equation}
\mathbf{M}_{[j]}^{(i)}
=
\gamma_{[j],C}^{(i)}
\overline{\mathbf{M}}_{[j]}^{(i)}.
\end{equation}
The cumulative transition over the first $j$ chunks of partition $i$ is therefore
\begin{equation}
\mathbf{M}_{[j\leftarrow1]}^{(i)}
=
\exp\left(
\sum_{r=1}^{j}
\log\gamma_{[r],C}^{(i)}
\right)
\overline{\mathbf{M}}_{[j\leftarrow1]}^{(i)},
\label{eq:appendix-gate-separated-transition}
\end{equation}
where
\begin{equation}
\overline{\mathbf{M}}_{[j\leftarrow1]}^{(i)}
=
\overline{\mathbf{M}}_{[j]}^{(i)}
\cdots
\overline{\mathbf{M}}_{[1]}^{(i)}.
\end{equation}

In direct low-precision multiplication, gate scaling is repeatedly incorporated into the cumulative transition, introducing additional rounding at each iteration. By separating the gate factors from the transition matrices, FlashQLA accumulates the gate-independent transitions without repeatedly applying gate scaling. The cumulative gate factor is tracked separately in the log domain and applied only when the full transition is needed.

\subsection{FP32-Accumulated Transition Updates}

The cumulative transition is computed through successive matrix multiplications across chunks. Directly accumulating the transition matrices in low precision introduces repeated rounding errors, which can compound over long partitions and degrade numerical accuracy.

To avoid this, we further simplify the cumulative transition update by expanding the chunk transition matrix. Using the gate-independent transition derived in Equation~\eqref{eq:gate-factored-transition}, we obtain
\begin{equation}
\begin{aligned}
\overline{\mathbf{M}}_{[j\leftarrow1]}^{(i)}
&=
\overline{\mathbf{M}}_{[j]}^{(i)}
\overline{\mathbf{M}}_{[j-1\leftarrow1]}^{(i)}
\\
&=
\left(
\mathbf{I}
-
\mathbf{K}_{[j]}^{(i)\top}
\mathbf{A}_{[j]}^{\prime(i)}
\mathbf{B}_{[j]}^{(i)}
\mathbf{K}_{[j]}^{(i)}
\right)
\overline{\mathbf{M}}_{[j-1\leftarrow1]}^{(i)}
\\
&=
\overline{\mathbf{M}}_{[j-1\leftarrow1]}^{(i)}
-
\mathbf{K}_{[j]}^{(i)\top}
\mathbf{A}_{[j]}^{\prime(i)}
\mathbf{B}_{[j]}^{(i)}
\mathbf{K}_{[j]}^{(i)}
\overline{\mathbf{M}}_{[j-1\leftarrow1]}^{(i)}.
\end{aligned}
\label{eq:appendix-fp32-transition}
\end{equation}

This reformulation separates the identity contribution from the matrix multiplication. FlashQLA maintains the cumulative transition in FP32 accumulators and computes the update term using low-precision matrix multiplications. Although the update term remains subject to low-precision arithmetic, the accumulated transition is preserved in FP32 across iterations, avoiding repeated low-precision rounding of the entire matrix.

\section{Derivation of Boundary-State Preprocessing}
\label{app:boundary-state-preprocessing}

We provide the derivation of the state-difference propagation property and boundary-state correction used in Section~\ref{sec:cp-theory}.

\paragraph{State-difference propagation.}
Consider two state trajectories $\mathbf{S}_{[c]}$ and $\widetilde{\mathbf{S}}_{[c]}$ that process identical chunk inputs but start from different initial states. By the affine chunk recurrence in Eq.~\eqref{eq:chunk-summary}, both trajectories satisfy
\begin{equation}
\begin{aligned}
\mathbf{S}_{[c+1]}
&=
\mathbf{M}_{[c]}\mathbf{S}_{[c]}
+
\mathbf{E}_{[c]},\\
\widetilde{\mathbf{S}}_{[c+1]}
&=
\mathbf{M}_{[c]}\widetilde{\mathbf{S}}_{[c]}
+
\mathbf{E}_{[c]}.
\end{aligned}
\end{equation}
Defining
$\Delta\mathbf{S}_{[c]}
=
\mathbf{S}_{[c]}-\widetilde{\mathbf{S}}_{[c]}$,
subtracting the two recurrences eliminates the identical additive contributions:
\begin{equation}
\Delta\mathbf{S}_{[c+1]}
=
\mathbf{M}_{[c]}\Delta\mathbf{S}_{[c]}.
\end{equation}
Repeatedly applying this relation over chunks $a$ through $b$ yields
\begin{equation}
\Delta\mathbf{S}_{[b+1]}
=
\mathbf{M}_{[b]}
\mathbf{M}_{[b-1]}
\cdots
\mathbf{M}_{[a]}
\Delta\mathbf{S}_{[a]},
\end{equation}
which establishes Eq.~\eqref{eq:bg-state-difference}.

\paragraph{Boundary-state correction.}
For partition $i$, let $\mathbf{S}_{[0]}^{(i)}$ denote its input boundary state. State-summary preparation performs a zero-state rollout initialized with
$\widetilde{\mathbf{S}}_{[0]}^{(i)}=\mathbf{0}$.
The true and zero-state trajectories follow the same chunk transitions, so their initial difference is
\begin{equation}
\Delta\mathbf{S}_{[0]}^{(i)}
=
\mathbf{S}_{[0]}^{(i)}.
\end{equation}
Applying the state-difference propagation property over all $l_i$ chunks gives
\begin{equation}
\mathbf{S}_{[l_i]}^{(i)}
-
\widetilde{\mathbf{S}}_{[l_i]}^{(i)}
=
\mathbf{M}_{[l_i\leftarrow1]}^{(i)}
\mathbf{S}_{[0]}^{(i)}.
\end{equation}
Rearranging yields
\begin{equation}
\mathbf{S}_{[l_i]}^{(i)}
=
\widetilde{\mathbf{S}}_{[l_i]}^{(i)}
+
\mathbf{M}_{[l_i\leftarrow1]}^{(i)}
\mathbf{S}_{[0]}^{(i)},
\end{equation}
recovering Eq.~\eqref{eq:bg-forward-boundary-correction}.

\section{Support for Zigzag Context Parallelism}
\label{app:zigzag-cp}

Modern LLM training frameworks often employ zigzag context parallelism to balance computation across GPUs in causal full-attention layers. This layout distributes non-contiguous sequence segments across GPUs, requiring GDN layers to support the same data layout for compatibility with hybrid attention architectures.

FlashQLA accommodates zigzag layouts through its hierarchical context-parallel design. At the inter-card level, each GPU computes state summaries and transitions for its local sequence segments. These summaries are then arranged according to the original sequence order, and inter-card boundary-state correction reconstructs the dependencies across segments, regardless of their physical GPU placement. The corrected boundary states are subsequently used to initialize the corresponding local computations, preserving the original GDN recurrence without requiring token-level redistribution.

Importantly, intra-card CP and Auto CP remain independent of the global zigzag layout. Once the boundary states of local segments are determined, intra-card CP operates on each segment as an ordinary contiguous sequence, while Auto CP selects its partitioning configuration based on the local workload. Neither mechanism needs to account for how segments are distributed across GPUs. This separation confines zigzag-specific handling to inter-card boundary-state correction and allows FlashQLA to reuse its existing intra-card execution and optimization mechanisms.

As zigzag adaptation only changes the organization and correction of segment-level summaries, it introduces minimal additional overhead without modifying the core computation kernels. In our evaluation, supporting zigzag layouts has no noticeable impact on overall training performance.

\end{document}